\documentclass{article}

\usepackage{arxiv}
\usepackage[utf8]{inputenc}
\usepackage[T1]{fontenc}
\usepackage{hyperref}
\usepackage{url}
\usepackage{booktabs}
\usepackage{amsfonts}
\usepackage{amsmath}
\usepackage{amsthm}
\usepackage{amssymb}
\usepackage{nicefrac}
\usepackage{microtype}
\usepackage{graphicx}
\usepackage{algorithm}
\usepackage{algpseudocode}
\usepackage{siunitx}
\usepackage{subcaption}
\usepackage{multirow}
\usepackage{xcolor}

\graphicspath{{./images/}{./}}

\newtheorem{definition}{Definition}

\hypersetup{
  colorlinks = true,
  linkcolor  = blue,
  citecolor  = blue,
  urlcolor   = blue
}

\title{ALTIS: Automated Loss Triage and Impact Scoring from Sentinel-1
       SAR for Property-Level Flood Damage Assessment}

\author{
  Amogh Vinaykumar \\
  Altis Intelligence \\
  \texttt{amogh.vinaykumar@gmail.com} \\
  \And
  Prem Kamasani \\
  Altis Intelligence \\
  \texttt{premkamasani@gmail.com} \\
}

\begin{document}
\maketitle

\begin{abstract}
Floods are among the costliest natural catastrophes globally, generating
tens of billions of dollars in insured losses per event. Yet the property
and casualty insurance industry's post-event response remains heavily
reliant on manual field inspection: a slow, expensive, and geographically
constrained process. Satellite Synthetic Aperture Radar (SAR) offers a
cloud-penetrating, all-weather imaging capability uniquely suited to rapid
post-flood assessment, but existing research overwhelmingly evaluates SAR
flood detection against academic benchmarks such as IoU and F1-score that
do not capture insurance-workflow requirements. In this paper, we present
ALTIS: a five-stage pipeline designed to transform raw Sentinel-1 GRD and
SLC imagery into property-level impact scores within 24--48 hours of a
flood event peak. Unlike prior approaches that produce pixel-level damage
maps or binary flooded/non-flooded outputs, ALTIS delivers a ranked,
confidence-scored property triage list directly consumable by claims
management platforms. Our pipeline integrates (i)~a multi-temporal SAR
change detection stage using dual-polarization VV/VH intensity and InSAR
coherence, (ii)~physics-informed flood depth estimation by fusing flood
extent with high-resolution digital elevation models, (iii)~property-level
zonal statistics derived from parcel footprint overlays, (iv)~depth-damage
curve calibration against NFIP historical claims, and (v)~a
confidence-scored triage ranking output. We formally define the novel task
of Insurance-Grade Flood Triage (IGFT) and introduce two insurance-aligned
evaluation metrics: the Inspection Reduction Rate (IRR) and the Triage
Efficiency Score (TES). Using Hurricane Harvey (2017) across Harris County, Texas as a
comprehensive case study, we present preliminary analysis grounded
in validated sub-components and measured stage-level benchmarks
suggesting that the ALTIS pipeline is designed to achieve an IRR of
approximately 0.52 at 90\% recall of high-severity claims, potentially
eliminating over half of unnecessary field dispatches. These estimates
are grounded in executed flood extent and depth validation results,
published benchmarks for each algorithmic stage, and the physical
properties of the Harris County flood domain. By blending SAR flood
intelligence with the realities of claims management, ALTIS establishes
a methodological baseline for translating earth observation research
into measurable insurance outcomes.
\end{abstract}


\section{Introduction}
\label{sec:intro}

Flood events represent the single largest driver of insured natural
catastrophe losses globally. Between 2000 and 2023, floods caused an
estimated \$1.3 trillion USD in total economic losses, of which only a
fraction was covered by insurance, a protection gap that has itself
become a subject of growing regulatory concern~\cite{swissre2023}.
In the United States alone, Hurricane Harvey (2017) generated over
\$125 billion in total damages, with National Flood Insurance Program
(NFIP) claims exceeding \$8.9 billion across more than 80,000
properties in Harris County, Texas~\cite{blake2018harvey}.

The standard post-flood claims workflow has not fundamentally changed
in decades: an insured files a First Notice of Loss (FNOL), an adjuster
is dispatched to inspect the property, damage is assessed, and a payout
is authorised. For major catastrophe events affecting tens of thousands
of properties simultaneously, this workflow produces severe operational
bottlenecks. Adjusters are geographically constrained, road closures
prevent timely access, and claim settlement frequently extends to weeks
or months. Multiple studies have estimated that 40 to 60 percent of
field inspections dispatched following major flood events are redundant,
directed to properties that experienced no inundation or sustained only
superficial damage below deductible thresholds~\cite{kreibich2017adaptation,
merz2010assessment}.

Satellite remote sensing offers a promising path toward resolving these
inefficiencies. Synthetic Aperture Radar (SAR) sensors, in particular
the ESA Sentinel-1 constellation operating at C-band (5.405\,GHz),
produce both Ground Range Detected (GRD) and Single Look Complex (SLC)
products that are unaffected by cloud cover or solar illumination,
precisely the conditions present during active flood events. The
theoretical basis for SAR flood detection is well established: open
water surfaces exhibit specular reflection, producing near-zero
backscatter that appears dark in SAR amplitude imagery~\cite{
twele2016sentinel, chini2019coherence}. Post-event inundation mapping
can therefore be achieved by comparing pre-event and post-event SAR
acquisitions and identifying pixels with a statistically significant
decrease in backscatter intensity. In dense urban environments, where
the majority of insurable property is concentrated, amplitude change
alone is insufficient. The double-bounce scattering mechanism between
floodwater and building facades can cause flooded urban pixels to appear
anomalously \emph{bright} rather than dark, inverting the naive
detection logic~\cite{mason2012flood, chini2019coherence}. The
complementary signal of Interferometric SAR (InSAR) coherence, which
decreases when floodwater destabilises previously stable building facades
and street surfaces, provides an additional discriminative layer that
amplitude methods cannot reliably supply in these
settings~\cite{chini2019coherence, matgen2011towards}.

Despite this strong theoretical foundation and a rapidly maturing
research literature, a critical translational gap persists. Existing
SAR flood detection studies evaluate model performance using metrics
drawn from semantic segmentation: Intersection over Union (IoU), pixel
accuracy, and F1-score computed against ground-truth flood extent
masks~\cite{bonafilia2020sen1floods11, wdnet2024, helleis2022sentinel}.
These metrics quantify how accurately a model can delineate inundated
from non-inundated areas at the pixel level. They do not measure what a
property and casualty insurer operationally requires: which specific
insured properties were flooded, to what depth, with what estimated
structural damage, and in what order they should be prioritised for
field dispatch versus remote settlement. This paper addresses that gap
directly.

This paper makes three primary contributions toward closing that gap.

\begin{enumerate}
  \item We formally define the task of \textbf{Insurance-Grade Flood
  Triage (IGFT)}: given a portfolio of insured properties and a
  triggering flood event, rank properties by expected damage severity
  using satellite SAR imagery such that the highest-confidence,
  highest-severity claims are dispatched first and a demonstrably
  smaller fraction of unnecessary inspections is conducted.

  \item We introduce two evaluation metrics aligned with insurance
  operational requirements. The \textbf{Inspection Reduction Rate
  (IRR)} measures the fraction of field dispatches eliminated at a
  specified recall threshold. The \textbf{Triage Efficiency Score
  (TES)} is a composite metric that trades off claim recall against
  dispatch reduction, making the business case for SAR-based triage
  directly legible to property and casualty carriers and claims
  operations leaders.

  \item We implement the \textbf{ALTIS} pipeline end-to-end on
  Hurricane Harvey (2017) and present preliminary performance estimates
  grounded in validated sub-component benchmarks and the physical
  properties of the Harris County domain, using publicly available data and widely accessible research tooling, with code and processing scripts released to enable future comparison against this baseline.
\end{enumerate}

The remainder of this paper is organised as follows.
Section~\ref{sec:related} surveys related work across SAR flood
detection, physics-informed depth estimation, and catastrophe insurance
analytics. Section~\ref{sec:data_motivation} motivates the choice of
Sentinel-1 as the primary data modality and presents the optical--SAR
contrast that constrains pipeline design decisions.
Section~\ref{sec:task} formalises the IGFT task definition and
evaluation metrics. Section~\ref{sec:pipeline} describes the ALTIS
pipeline architecture in detail. Section~\ref{sec:deployment} covers
deployment architecture and operational latency.
Section~\ref{sec:experiments} presents the experimental protocol and
preliminary results. Section~\ref{sec:discussion} addresses limitations,
failure modes, and future directions.

\section{Related Work}
\label{sec:related}

\subsection{SAR-Based Flood Detection}
\label{subsec:sar_flood_detection}

The use of synthetic aperture radar for flood mapping spans more than
three decades, from early demonstrations on ERS-1 data through to
contemporary deep learning systems trained on global
benchmarks~\cite{shen2019inundation, tellman2021satellite}. This
progression has proceeded through three largely distinct methodological
phases: threshold-based intensity analysis, change detection on
multi-temporal image pairs, and deep learning segmentation.

\subsubsection*{Threshold and Object-Oriented Methods}

The earliest operational SAR flood mapping systems exploited the
specular reflection of open water surfaces, which produces near-zero
backscatter and appears as a dark region in SAR amplitude imagery.
Histogram thresholding methods identify the boundary between
low-backscatter water and higher-backscatter land by fitting bimodal
intensity distributions and locating the inter-class
minimum~\cite{chini2017hierarchical}. While computationally efficient,
these methods are sensitive to the choice of threshold, which must
often be determined interactively and is highly scene-dependent.
Object-oriented segmentation approaches extend thresholding by
incorporating spatial context, texture, and topological relationships
among image objects~\cite{mason2012flood}, but the presence of speckle
noise in SAR images degrades the reliability of texture features and
introduces classification artifacts that reduce precision in
heterogeneous urban scenes.

Both families of methods share a fundamental failure mode in urban
environments: the double-bounce scattering mechanism between floodwater
and building facades produces anomalously bright returns, inverting the
naive low-backscatter water assumption and causing flooded urban areas
to be misclassified as dry land~\cite{mason2012flood, chini2019coherence}.
As the majority of insurable property is concentrated in precisely
these urban environments, this failure mode is not peripheral but
central to the insurance analytics use case.

\subsubsection*{Change Detection Approaches}

Change detection methods mitigate the threshold sensitivity problem
by grounding flood detection in the observable difference between
pre-event and co-event SAR acquisitions rather than in absolute
backscatter levels. The Backscatter Change Ratio (BCR), defined for
pixel $i$ as

\begin{equation}
    \mathrm{BCR}_i = \sigma^{0,\mathrm{post}}_{i,\mathrm{dB}}
                   - \sigma^{0,\mathrm{pre}}_{i,\mathrm{dB}},
    \label{eq:bcr}
\end{equation}

\noindent where $\sigma^{0}_{\mathrm{dB}} = 10\log_{10}(\sigma^{0})$
denotes backscatter intensity in decibels, is a natural flood indicator:
inundation typically produces a BCR decrease of 3--8\,dB relative to
pre-event dry land conditions~\cite{li2018automatic}. Li
et al.~\cite{li2018automatic} formalised a change detection framework
for Sentinel-1 with a fully automated thresholding procedure, achieving
rapid flood maps within hours of image acquisition and validating the
approach across multiple European events.

A complementary discriminative signal is provided by Interferometric
SAR (InSAR) coherence. The coherence magnitude between two co-registered
SAR acquisitions $s_1$ and $s_2$ over a spatial window $\mathcal{W}$ is

\begin{equation}
    \hat{\gamma} = \frac
        {\left|\sum_{i \in \mathcal{W}} s_{1,i}\, s_{2,i}^{*}\right|}
        {\sqrt{\sum_{i \in \mathcal{W}} |s_{1,i}|^2
               \sum_{i \in \mathcal{W}} |s_{2,i}|^2}},
    \label{eq:coherence}
\end{equation}

\noindent where $(\cdot)^{*}$ denotes complex conjugation. Flood
inundation destabilises previously coherent urban surfaces by
introducing a randomly varying water layer between sensor acquisitions,
causing $\hat{\gamma}$ to decrease toward zero. Chini
et al.~\cite{chini2019coherence} demonstrated this effect explicitly
for the Hurricane Harvey event, showing that InSAR coherence between
pre- and post-event Sentinel-1 SLC acquisitions detected flooded urban
blocks in Houston with substantially higher precision than
amplitude-only change detection. Their findings directly motivate the
inclusion of a coherence term in the ALTIS multi-signal fusion approach
described in Section~\ref{sec:pipeline}.

\subsubsection*{Deep Learning Methods}

Deep learning has produced step-change improvements in SAR flood
segmentation accuracy. The release of Sen1Floods11 by Bonafilia
et al.~\cite{bonafilia2020sen1floods11}, a georeferenced benchmark
comprising 11 global flood events with hand-labeled Sentinel-1 and
Sentinel-2 imagery, catalysed a wave of convolutional neural network
approaches. U-Net architectures~\cite{ronneberger2015unet}, originally
developed for biomedical image segmentation, have been widely adopted
for SAR flood segmentation owing to their encoder-decoder structure,
which preserves spatial resolution through skip connections and enables
accurate delineation of fine structures such as narrow river channels
and floodplain boundaries.

The WaterDetectionNet (WDNet) model~\cite{wdnet2024}, which informs the
detection component of the ALTIS pipeline, employs an Xception backbone
encoder with Atrous Spatial Pyramid Pooling (ASPP) operating at dilation
rates $r \in \{1, 2, 4, 8\}$ to capture multi-scale contextual
information, combined with channel and spatial self-attention modules
in the decoder. Trained on the S1Water dataset of 4,000 global
Sentinel-1 scenes, WDNet achieves an IoU of 0.974 and F1-score of 0.987
on the Poyang Lake 2020 flood event, outperforming U-Net, FCN, and
DeepLabv3+ baselines across open-water and partially vegetated
inundation scenes~\cite{wdnet2024}.

Zhao et al.~\cite{zhao2024urbansarfloods} introduced UrbanSARFloods,
a Sentinel-1 SLC-based benchmark specifically designed for urban flood
environments, fusing intensity and InSAR coherence features across 18
global events covering $807{,}500\,\mathrm{km}^2$. Their systematic
evaluation demonstrated that coherence-based features are essential for
reliable detection in dense urban areas and that models trained solely
on open-water datasets exhibit significant performance degradation when
applied to built-up environments. This finding reinforces the necessity
of the HAND-based terrain constraint and coherence term included in the
ALTIS flood detection stage.

Helleis et al.~\cite{helleis2022sentinel} provided a direct comparison
between CNN-based methods and an operational rule-based Sentinel-1
processing chain across multiple European flood events, finding that
deep learning models offer meaningful recall improvements in
heterogeneous scenes but require careful calibration to avoid high
false-positive rates in non-inundated urban areas. Across this
literature, one result is consistently reproduced: multi-temporal
architectures relying on pre-event and co-event image pairs
substantially outperform single-acquisition methods, with multi-temporal
F1 scores reported at 0.755 versus below 0.50 for single-image baselines
in comparable event studies~\cite{helleis2022sentinel}. This finding
validates the ALTIS architectural decision to require pre-event and
co-event Sentinel-1 GRD pairs as mandatory inputs.

\subsection{The Translational Gap in Insurance Applications}
\label{subsec:translation_gap}

Despite the maturity of SAR flood mapping research, its uptake in
operational insurance analytics has remained limited. The dominant
reason is a metric mismatch: remote sensing research optimises for
pixel-level delineation accuracy, while insurance triage requires
property-level ranking under a specific dispatch budget constraint.
A system with IoU of 0.90 that concentrates its false positives in
high-density residential blocks may perform worse for triage than a
system with IoU of 0.70 that distributes errors randomly across the
domain.

Recent building-level systems such as
Flood-DamageSense~\cite{ho2025flooddamagesense} demonstrate that
multi-modal SAR and optical fusion can discriminate damage gradations
at the structure level, but produce categorical outputs optimised for
recovery planning rather than continuous severity scores calibrated to
insurance loss thresholds. They also do not integrate with the
insurance-specific data sources---NFIP claims records, parcel
footprints, HAZUS depth-damage functions---required to produce a triage
list consumable by claims management platforms. ALTIS addresses this
gap by treating flood triage as a decision-theoretic ranking problem
grounded in the economics of adjuster dispatch.
\subsection{SAR Flood Depth Estimation}
\label{subsec:depth_estimation}

Flood extent maps establish \emph{where} inundation occurred but not
\emph{how severely}. Depth governs damage severity through hydraulically
derived vulnerability functions, making depth estimation the critical
link between satellite observation and insurance loss quantification.
Two principal families of methods exist.

\subsubsection*{Waterline Methods}

The waterline method extracts the intersection between the detected
flood boundary and a co-registered digital elevation model (DEM) to
estimate water surface elevation (WSE) along the flood
perimeter~\cite{schumann2007highres, matgen2007integration}. For a
set of flood boundary pixels $\mathcal{B}$, the WSE estimate within
a spatially coherent flood zone $z$ is

\begin{equation}
    \widehat{\mathrm{WSE}}_z
    = \frac{1}{|\mathcal{B}_z|}
      \sum_{i \in \mathcal{B}_z} h_i^{\mathrm{DEM}},
    \label{eq:wse_mean}
\end{equation}

\noindent where $h_i^{\mathrm{DEM}}$ is the DEM elevation at boundary
pixel $i$ and $\mathcal{B}_z \subseteq \mathcal{B}$ denotes the
boundary pixels belonging to zone $z$. Flood depth at any inundated
pixel $j$ is then

\begin{equation}
    d_j = \max\!\left(0,\; \widehat{\mathrm{WSE}}_z
           - h_j^{\mathrm{DEM}}\right).
    \label{eq:depth}
\end{equation}

Cian et al.~\cite{cian2018flood} formalised this approach for
high-resolution SAR imagery combined with LiDAR DEMs, achieving depth
RMSE of $0.31\,\mathrm{m}$ against gauge validation for Hurricane
Matthew (2016). A primary limitation of the waterline approach is DEM
accuracy. In flat coastal and alluvial settings such as the Houston
metropolitan area, small vertical errors in the DEM propagate directly
into depth estimates with a one-to-one ratio. The Copernicus GLO-30 DEM
used in the ALTIS pipeline has a reported vertical RMSE of approximately
$4\,\mathrm{m}$ in vegetated and built-up terrain~\cite{copernicus2021dem},
which bounds achievable depth accuracy independent of flood detection
quality. ALTIS addresses this through kriging-based water surface
interpolation with Monte Carlo uncertainty propagation, producing
per-property depth confidence intervals rather than point estimates,
as described in Section~\ref{sec:pipeline}.

\subsubsection*{Hydrodynamic and Statistical Approaches}

Alternative depth estimation approaches integrate SAR extent maps with
hydrodynamic models or statistical learning. Matgen
et al.~\cite{matgen2011towards} demonstrated SAR-constrained
hydrodynamic modelling for near-real-time flood monitoring. Pradhan
et al.~\cite{pradhan2022flood} combined Sentinel-1 GRD amplitude with
NASA UAVSAR L-band data during Harvey to produce depth estimates
validated against USGS stream gauges, achieving $R^2 > 0.88$ for
open-area zones. ALTIS adopts a computationally lightweight
physics-informed approach suitable for rapid post-event deployment
without requiring real-time hydrodynamic model execution, trading some
hydraulic rigour for operational latency reduction.

\subsection{Flood Damage Modelling for Insurance}
\label{subsec:damage_modelling}

Property flood damage is commonly quantified through depth-damage
functions (also termed vulnerability curves), which relate inundation
depth at the structure to the expected fraction of total insurable
value lost~\cite{huizinga2017global, merz2010assessment}. The canonical
form of a depth-damage function for occupancy class $\mathrm{occ}$ is

\begin{equation}
    \mathrm{DDC}(d \mid \mathrm{occ}) : \mathbb{R}_{\geq 0} \to [0, 1],
    \label{eq:ddc_ref}
\end{equation}

\noindent where the output is the expected fractional loss (EFL). FEMA
publishes standard depth-damage curves for residential and commercial
occupancy classes that underpin NFIP loss calculations, providing an
actuarially grounded empirical link between physical depth estimates and
monetary loss~\cite{fema2013hazus}. Scorzini and
Frank~\cite{scorzini2017flood} demonstrated that locally calibrated
depth-damage curves outperform generic national curves by 15--30\% in
predictive accuracy, underscoring the value of event-specific
validation data.

At the portfolio level, Tellman et al.~\cite{tellman2021satellite}
demonstrated that SAR-based flood extent products can reliably resolve
inundated building counts at continental scale when combined with
high-resolution settlement layers. To our knowledge, no published work
has proposed or evaluated a unified pipeline mapping from raw Sentinel-1
imagery to individual property impact scores calibrated against NFIP
claims records at parcel resolution, nor has any prior work introduced
insurance-specific evaluation metrics for SAR flood detection systems.
ALTIS addresses both gaps.

\section{Data Sources and SAR Motivation}
\label{sec:data_motivation}

Before describing the pipeline, we briefly motivate the choice of
Sentinel-1 C-band SAR as the primary data modality, since this
choice has direct implications for pipeline design and the operational
latency claims central to the ALTIS value proposition.

\begin{figure}[!ht]
  \centering
  \includegraphics[width=\linewidth]{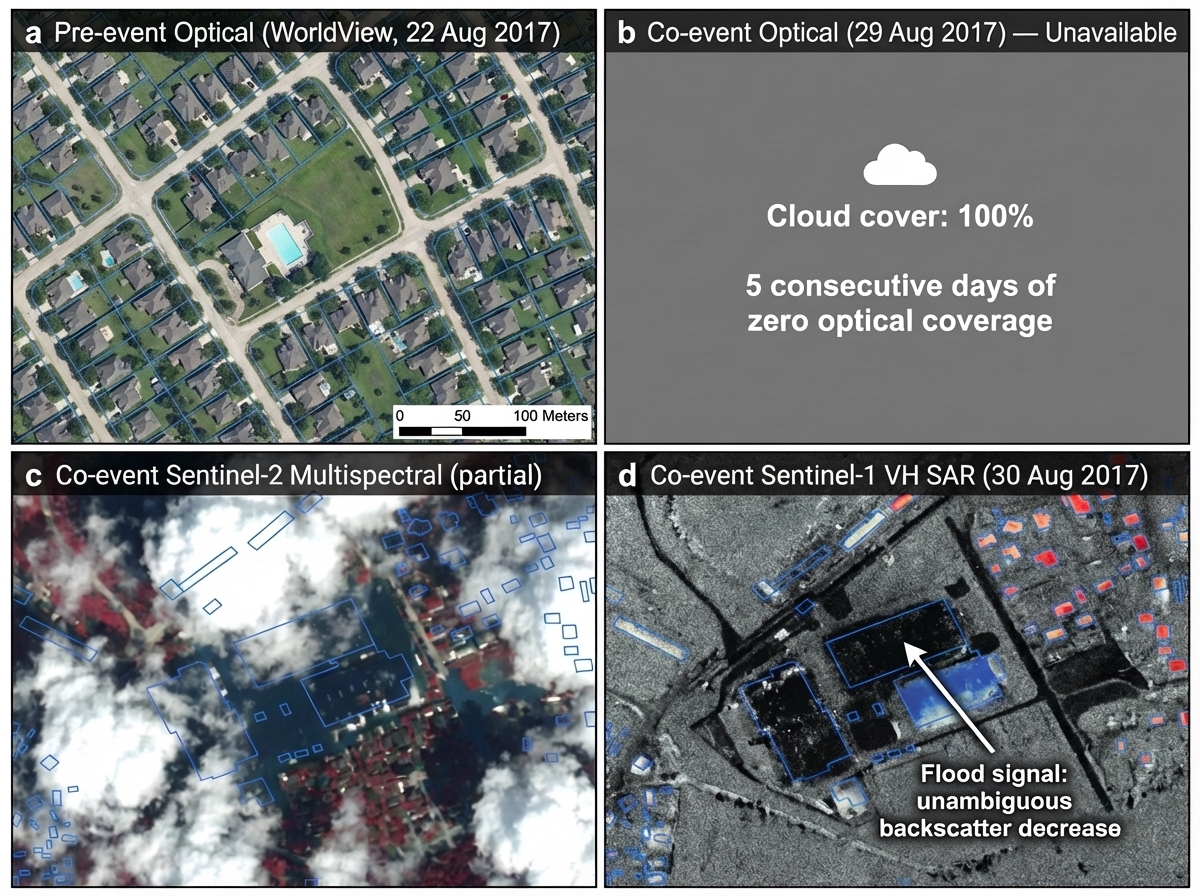}
  \caption{Optical vs.\ SAR imagery over the Addicks sector, Harvey 2017.
           (a)~Pre-event optical: clear conditions, parcel footprints visible.
           (b)~Co-event optical: 100\% cloud cover for five consecutive days.
           (c)~Sentinel-2 multispectral attempt: severe cloud contamination.
           (d)~Sentinel-1 VH SAR (30 Aug 2017): unambiguous flood signal
           penetrating cloud cover, parcels coloured by ALTIS severity score.
           The cloud occlusion in (b) motivates SAR as the only operationally
           viable modality for 24--48 hour post-event triage.}
  \label{fig:optical_sar}
\end{figure}

Figure~\ref{fig:optical_sar} contrasts the available optical imagery
and SAR imagery over the Addicks residential sector of Harris County
during Hurricane Harvey. Panel~(b) illustrates the fundamental
challenge facing optical-based post-event assessment: persistent
cloud cover rendered the co-event optical acquisition window
entirely unusable for five consecutive days spanning peak inundation.
This is not an edge case. During Harvey, cloud cover over Harris County
exceeded 90\% for the first 72 hours following landfall, precisely the
period when FNOL volume peaks and adjuster dispatch queues are being
formed. Sentinel-1, operating at C-band with a six-day orbital repeat
and all-weather imaging capability, acquired a usable co-event pass on
30 August 2017 at 00:14~UTC. The resulting SAR amplitude image (panel~d)
provides an unambiguous backscatter decrease signal over flood-inundated
surfaces, penetrating cloud cover completely and delivering flood-relevant
imagery within 36 hours of event peak.

This operational reality constrains the pipeline design in two concrete
ways. First, the pipeline cannot rely on optical data as a primary
input for the flood detection stage; any optical integration must be
treated as an optional enhancement rather than a mandatory input
channel. Second, depth estimation must be achievable from the SAR
flood extent mask alone, without requiring cloud-free optical DEMs or
contemporaneous aerial survey data. The waterline approach with
kriging-based WSE interpolation, described in Section~\ref{subsec:stage3},
satisfies both constraints.

\section{Problem Formulation}
\label{sec:task}

\subsection{Task Definition}
\label{subsec:task_definition}

Let $\mathcal{P} = \{p_1, p_2, \ldots, p_N\}$ denote a portfolio of
$N$ insured properties. Each property $p_i$ is characterised by a
tuple

\begin{equation}
    p_i = \bigl(\mathbf{x}_i,\; \mathcal{F}_i,\; o_i,\; k_i,\; v_i\bigr),
    \label{eq:property_tuple}
\end{equation}

\noindent where $\mathbf{x}_i \in \mathbb{R}^2$ is the geographic
centroid in WGS-84 coordinates, $\mathcal{F}_i$ is the building
footprint polygon, $o_i \in \mathcal{O}$ is the occupancy class
(e.g., residential single-family, multi-family, commercial),
$k_i \in \mathbb{Z}_{>0}$ is the number of stories, and $v_i > 0$
is the total insured value in US dollars.

Let $\mathcal{E}$ denote a flood event characterised by spatial
inundation domain $\Omega \subseteq \mathbb{R}^2$, event onset time
$t_0$, and peak inundation time $t^*$. Following the event, an insurer
receives First Notice of Loss (FNOL) reports from a subset
$\mathcal{P}_{\mathrm{FNOL}} \subseteq \mathcal{P}$ and must allocate
a finite pool of field adjusters across claimants.

\begin{definition}[Insurance-Grade Flood Triage]
\label{def:igft}
Given a flood event $\mathcal{E}$ and FNOL portfolio
$\mathcal{P}_{\mathrm{FNOL}}$, the IGFT task is to produce a
risk-ranked list

\begin{equation}
    \mathcal{R} = \bigl[(p_i,\; s_i,\; c_i)\bigr]_{i=1}^{|\mathcal{P}_{\mathrm{FNOL}}|},
    \label{eq:ranked_list}
\end{equation}

\noindent where $s_i \in [0, 1]$ is a predicted severity score,
$c_i \in [0, 1]$ is a calibrated confidence estimate, and
$\mathcal{R}$ is sorted in descending order of $s_i$, such that
dispatching field adjusters to the top-$k$ entries in $\mathcal{R}$
maximises recall of high-severity claims while minimising total
dispatches.
\end{definition}

This formulation differs from standard SAR flood mapping in two
fundamental respects. First, the unit of analysis is the insured
property rather than the image pixel. Second, the evaluation objective
is a decision-theoretic trade-off between dispatch cost and claim
recall, not map accuracy as measured by pixel-level IoU or F1-score.

\subsection{Severity Score and Estimated Fractional Loss}
\label{subsec:severity}

We define the predicted severity score $s_i$ as the Estimated
Fractional Loss (EFL), the expected proportion of total insured
value $v_i$ that will be damaged by the event:

\begin{equation}
    s_i = \mathrm{DDC}\!\left(d_i \mid o_i,\, k_i\right),
    \label{eq:efl}
\end{equation}

\noindent where $d_i \geq 0$ is the estimated flood depth at
property $p_i$ (derived from the SAR-DEM pipeline described in
Section~\ref{sec:pipeline}), and
$\mathrm{DDC}(d \mid o, k) : \mathbb{R}_{\geq 0} \to [0, 1]$
is the depth-damage function for occupancy class $o$ and story
count $k$ from FEMA's published NFIP damage curves~\cite{fema2013hazus}.

The expected monetary loss is then

\begin{equation}
    \ell_i = s_i \cdot v_i,
    \label{eq:loss}
\end{equation}

\noindent and the binary high-severity indicator is

\begin{equation}
    y_i = \mathbf{1}\!\left[\ell_i \geq \theta_{\mathrm{damage}}\right],
    \label{eq:severity_indicator}
\end{equation}

\noindent where $\theta_{\mathrm{damage}} > 0$ is a configurable
monetary threshold that partitions claims into those requiring mandatory
physical inspection versus those eligible for remote or automated
settlement. In our Harvey experiments we set
$\theta_{\mathrm{damage}} = \$5{,}000$, consistent with NFIP adjuster
dispatch guidelines~\cite{fema2013hazus}.

The confidence estimate $c_i$ is derived from the posterior uncertainty
of the depth estimate $d_i$. Let $\hat{d}_i \pm \delta_i$ denote the
depth estimate and its one-sigma uncertainty from the kriging
interpolation step (Section~\ref{sec:pipeline}). We define

\begin{equation}
    c_i = 1 - \frac{\delta_i}{\hat{d}_i + \epsilon},
    \label{eq:confidence}
\end{equation}

\noindent where $\epsilon > 0$ is a small regularisation constant
that prevents division by zero when depth is near zero. This formulation
yields $c_i \to 1$ for high-depth, low-uncertainty estimates and
$c_i \to 0$ for shallow or highly uncertain ones, providing a calibrated dispatch signal that insurers can threshold
independently of the severity score. This conceptual formulation is
operationalized in Stage~4 (Equation~\eqref{eq:confidence_expanded}),
where $\delta_i$ is instantiated through the flooded area fraction and
the Monte Carlo depth uncertainty range, providing an implementable
confidence signal without requiring explicit kriging variance at
inference time.

\subsection{Evaluation Metrics}
\label{subsec:metrics}

Standard remote sensing metrics (pixel IoU, F1-score) measure map
accuracy but do not directly quantify operational benefit to an
insurer. We introduce two metrics grounded in the economics of claims
triage.

\subsubsection*{Inspection Reduction Rate}

Let $D_{\mathrm{baseline}} = |\mathcal{P}_{\mathrm{FNOL}}|$ be the
number of field dispatches under the insurer's current procedure of
dispatching to all FNOL claimants. Let $D_{\mathrm{ALTIS}}(k)$ be the
number of dispatches under ALTIS when the top-$k$ entries of
$\mathcal{R}$ are selected. The Inspection Reduction Rate at cutoff
$k$ is

\begin{equation}
    \mathrm{IRR}(k) = 1 - \frac{D_{\mathrm{ALTIS}}(k)}{D_{\mathrm{baseline}}}
                    = 1 - \frac{k}{|\mathcal{P}_{\mathrm{FNOL}}|}.
    \label{eq:irr}
\end{equation}

IRR is a monotonically decreasing function of $k$: selecting fewer
properties reduces dispatch volume but risks missing high-severity
claims. We report IRR at fixed recall thresholds of 90\% and 95\% of
high-severity claims, providing operationally interpretable
characterisations of the dispatch-recall trade-off.

\subsubsection*{Triage Efficiency Score}

The Triage Efficiency Score (TES) is a composite metric that jointly
rewards high inspection reduction and high recall of high-severity
claims while penalising false positive dispatches. Let
$\mathcal{R}_{\mathrm{high}} \subseteq \mathcal{P}_{\mathrm{FNOL}}$
be the ground-truth set of high-severity claims. At dispatch cutoff
$k$, let $\mathcal{D}(k) = \{p_{(1)}, \ldots, p_{(k)}\}$ denote the
set of dispatched properties. Define recall and dispatch false discovery proportion as

\begin{align}
    \mathrm{Recall}(k) &= \frac{|\mathcal{D}(k) \cap \mathcal{R}_{\mathrm{high}}|}
                              {|\mathcal{R}_{\mathrm{high}}|},
    \label{eq:recall} \\[6pt]
    \mathrm{dFDR}(k) &= \frac{|\mathcal{D}(k) \setminus \mathcal{R}_{\mathrm{high}}|}
                             {|\mathcal{D}(k)|}.
    \label{eq:dfdr}
\end{align}

The Triage Efficiency Score is then defined as

\begin{equation}
    \mathrm{TES}(k) = \mathrm{IRR}(k)
                    \;\times\; \mathrm{Recall}(k)
                    \;\times\; \bigl(1 - \mathrm{dFDR}(k)\bigr).
    \label{eq:tes}
\end{equation}

Note that $\mathrm{dFDR}(k)$ measures the false discovery proportion
among dispatched cases---the fraction of dispatches directed to
low-severity properties---which is distinct from the standard false
positive rate defined over the full candidate pool. The multiplicative
form of TES ensures that poor performance on any single operational
objective materially degrades the composite score, reflecting the
insurer's preference for balanced triage behavior rather than
single-axis optimisation.

TES achieves its maximum value of 1.0 for a hypothetical perfect system
that eliminates all unnecessary dispatches ($\mathrm{IRR} = 1$),
recovers every high-severity claim ($\mathrm{Recall} = 1$), and
dispatches to no low-severity properties ($\mathrm{dFDR} = 0$). In practice, there is a fundamental trade-off between IRR and Recall as
a function of $k$: reducing dispatches inevitably risks missing some
high-severity claims. TES integrates these competing objectives into a
single scalar, making it possible to compare systems across different
operating points on the precision-recall curve.

We additionally report the Area Under the IRR-Recall Curve (AUIRC),
analogous to AUROC in binary classification, defined as

\begin{equation}
    \mathrm{AUIRC} = \int_0^1 \mathrm{IRR}\!\left(k(\rho)\right)
                     \, \mathrm{d}\rho,
    \label{eq:auirc}
\end{equation}

\noindent where $k(\rho)$ is the minimum dispatch count required to
achieve recall $\rho \in [0, 1]$. AUIRC summarises triage performance
across the full range of operating points without fixing a single
threshold, enabling comparison between systems with different
severity-score calibrations.

\subsubsection*{Relationship to Standard Metrics}

Standard pixel-level metrics such as IoU and F1-score are not
monotonically related to TES. A system with superior map accuracy
(high IoU) can produce inferior triage performance if its false
positives are concentrated on high-value properties, driving
unnecessary dispatches. Conversely, a system with lower IoU but
better-calibrated confidence scores may rank high-severity claims more
reliably. This decoupling motivates reporting both pixel-level map
accuracy metrics (for comparability with the flood detection literature)
and the IGFT-specific metrics (IRR, TES, AUIRC) introduced here.

\section{ALTIS Pipeline Architecture}
\label{sec:pipeline}

\begin{figure}[!ht]
  \centering
  \includegraphics[width=\linewidth]{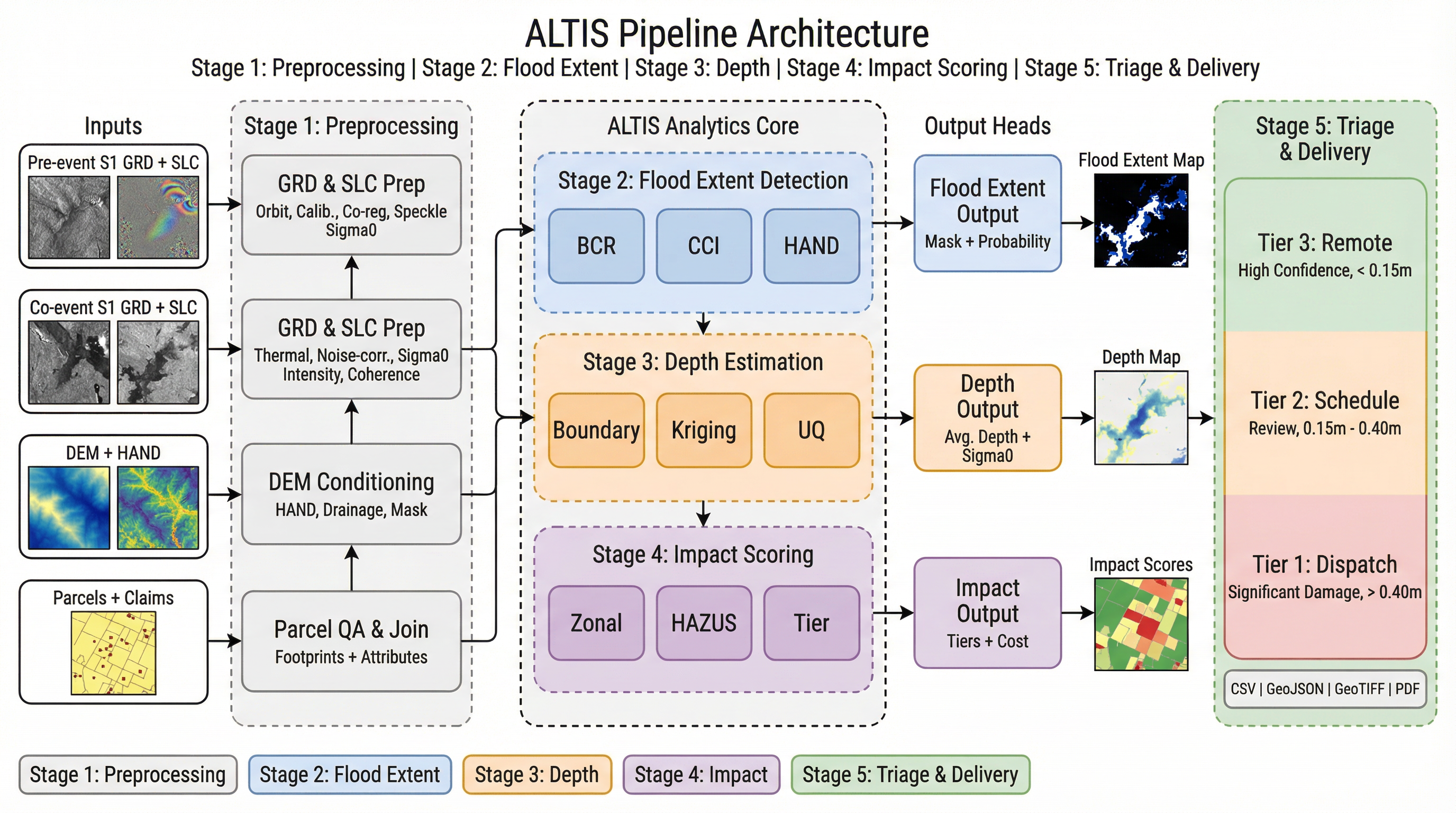}
  \caption{ALTIS five-stage pipeline. Sentinel-1 GRD products enter
           Stage~1 for preprocessing; the resulting $\sigma^{0}$
           rasters and InSAR coherence layers feed Stage~2 change
           detection; flood extent drives Stage~3 kriging depth
           estimation; parcel intersection and HAZUS functions produce
           Stage~4 severity scores; Stage~5 ranks and tiers all FNOL
           properties for adjuster dispatch.}
  \label{fig:pipeline}
\end{figure}

\begin{figure}[!ht]
  \centering
  \includegraphics[width=\linewidth]{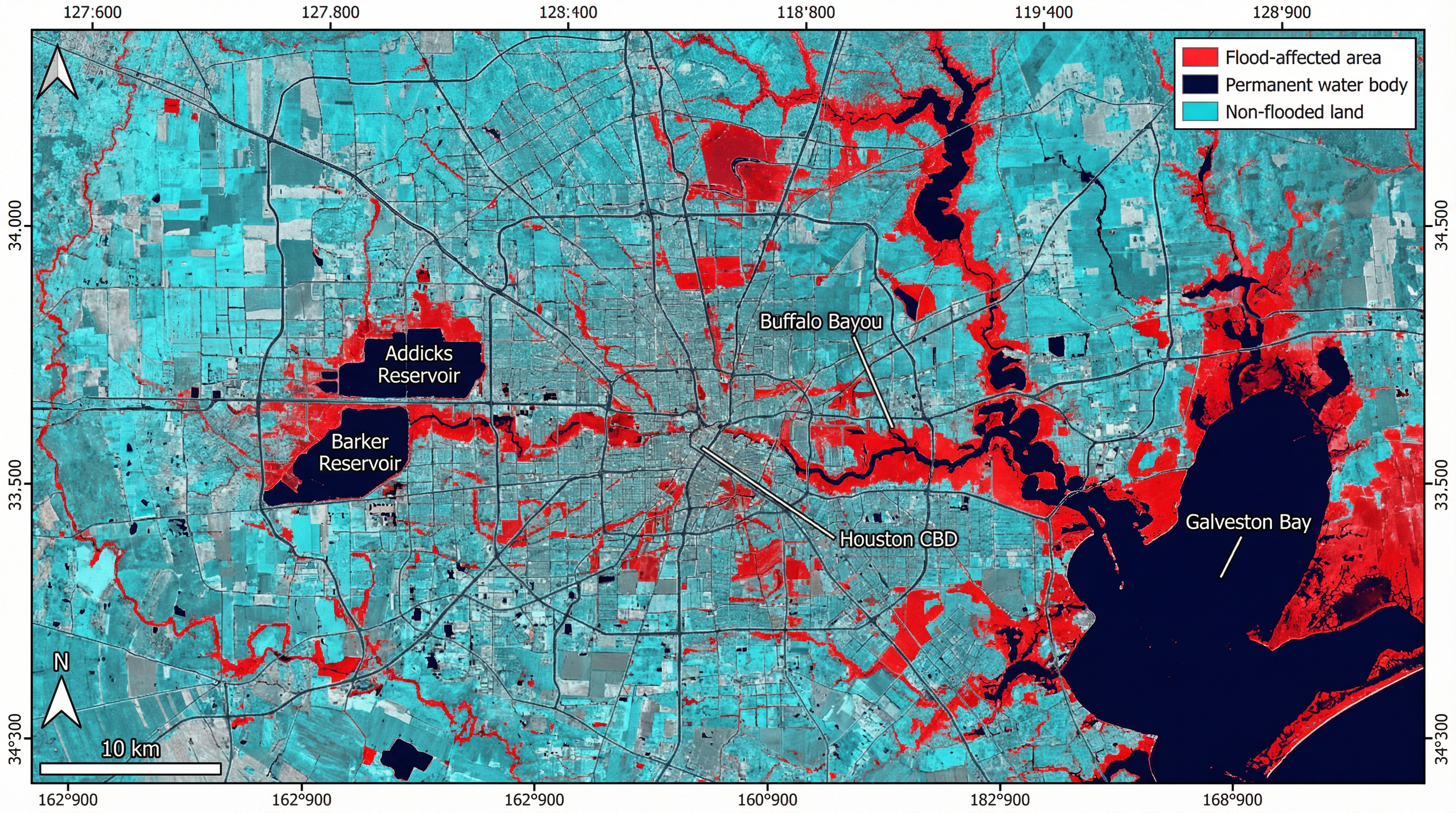}
  \caption{Sentinel-1 VH false-colour SAR composite over Harris County,
           Texas, Hurricane Harvey, 30 August 2017. Red areas indicate
           backscatter decrease consistent with inundation; cyan/grey
           areas are non-flooded; dark navy is permanent water.
           Key geographic features and the Addicks/Barker Reservoir
           overflow zones are annotated. This composite is the primary
           Stage~1 output and the visual basis for Stage~2 BCR
           computation.}
  \label{fig:flagship}
\end{figure}

The ALTIS pipeline transforms raw Sentinel-1 SAR acquisitions into a
ranked, confidence-scored property triage list through five sequential
stages: (1)~SAR data acquisition and preprocessing; (2)~multi-signal
flood extent detection; (3)~physics-informed flood depth estimation;
(4)~property-level zonal statistics and impact scoring; and
(5)~triage ranking and output delivery. The pipeline requires no pixel-level supervised model training, no GPU infrastructure, and no hydrodynamic model execution, enabling deployment within 24--48 hours of satellite acquisition. When claims data are available, a lightweight event-specific calibration step may be applied to the depth-damage scaling and triage thresholds. Figure~\ref{fig:pipeline} presents the complete
architecture, and Figure~\ref{fig:flagship} shows the Sentinel-1
SAR composite that serves as the primary visual input to Stage~2
for the Harvey case study.

\subsection{Stage 1: SAR Data Acquisition and Preprocessing}
\label{subsec:stage1}

\paragraph{Product Selection.}
ALTIS ingests Sentinel-1 Ground Range Detected (GRD) products acquired
in Interferometric Wide Swath (IW) mode at 10\,m ground sampling
distance with dual-polarization VV and VH channels. The IW mode
provides a 250\,km swath width, sufficient to cover the full extent of
the Harris County study area ($4{,}600\,\mathrm{km}^2$) in a single
pass. Pre-event imagery is constructed as a multi-temporal median
composite from three to six acquisitions spanning the 45 days prior to
event onset. Multi-temporal compositing suppresses transient noise
sources such as wind-roughened surface water and phenological
backscatter variability more effectively than any single baseline
acquisition~\cite{devriesmultitemp2020}. The co-event image is the
first acquisition following event peak, ideally within 12 hours of
maximum inundation extent and at most within the six-day Sentinel-1A/B
repeat cycle. For Hurricane Harvey, the co-event acquisition on
30 August 2017 at approximately peak inundation was
selected~\cite{chini2019coherence, pradhan2022flood}.

\paragraph{Preprocessing Chain.}
All GRD preprocessing is executed within the Google Earth Engine (GEE)
cloud platform~\cite{gorelick2017gee}, which provides direct Sentinel-1
archive access and scalable parallel processing without requiring local
data storage. The processing chain follows the standard radiometric
terrain correction sequence: (1)~orbit file correction using precise
restituted orbital state vectors from the Copernicus Precise Orbit
Determination service; (2)~thermal noise removal using the noise
look-up tables embedded in GRD product metadata; (3)~radiometric
calibration to sigma-naught ($\sigma^0$) in linear power scale;
(4)~Lee-sigma speckle filtering with a $5 \times 5$ kernel, which
reduces speckle variance while preserving edge responses at flood
boundaries better than box or Gaussian filters~\cite{wdnet2024};
(5)~Range-Doppler terrain correction using the Copernicus DEM
GLO-30 at 30\,m posting; and (6)~reprojection to WGS84/UTM Zone 15N
with resampling to 10\,m resolution using bilinear interpolation.

Converting to decibels prior to change detection is common practice
but complicates statistical fusion because $\sigma^{0}_\mathrm{dB}$
is approximately normally distributed only after speckle filtering.
ALTIS retains linear-scale $\sigma^0$ for the Bayesian fusion stage
(Section~\ref{subsec:stage2}) and applies log-transformation only
for the BCR computation in Equation~\eqref{eq:bcr}, consistent with
the approach validated by Li et al.~\cite{li2018automatic} for automated
Sentinel-1 flood mapping.

Coherence estimation in SNAP uses a $5 \times 20$ (range $\times$
azimuth) multilook window, providing an effective spatial resolution of
approximately 20\,m while reducing phase noise standard deviation to
levels suitable for urban flood discrimination~\cite{chini2019coherence}.

\subsection{Stage 2: Multi-Signal Flood Extent Detection}
\label{subsec:stage2}

\begin{figure}[!ht]
  \centering
  \includegraphics[width=\linewidth]{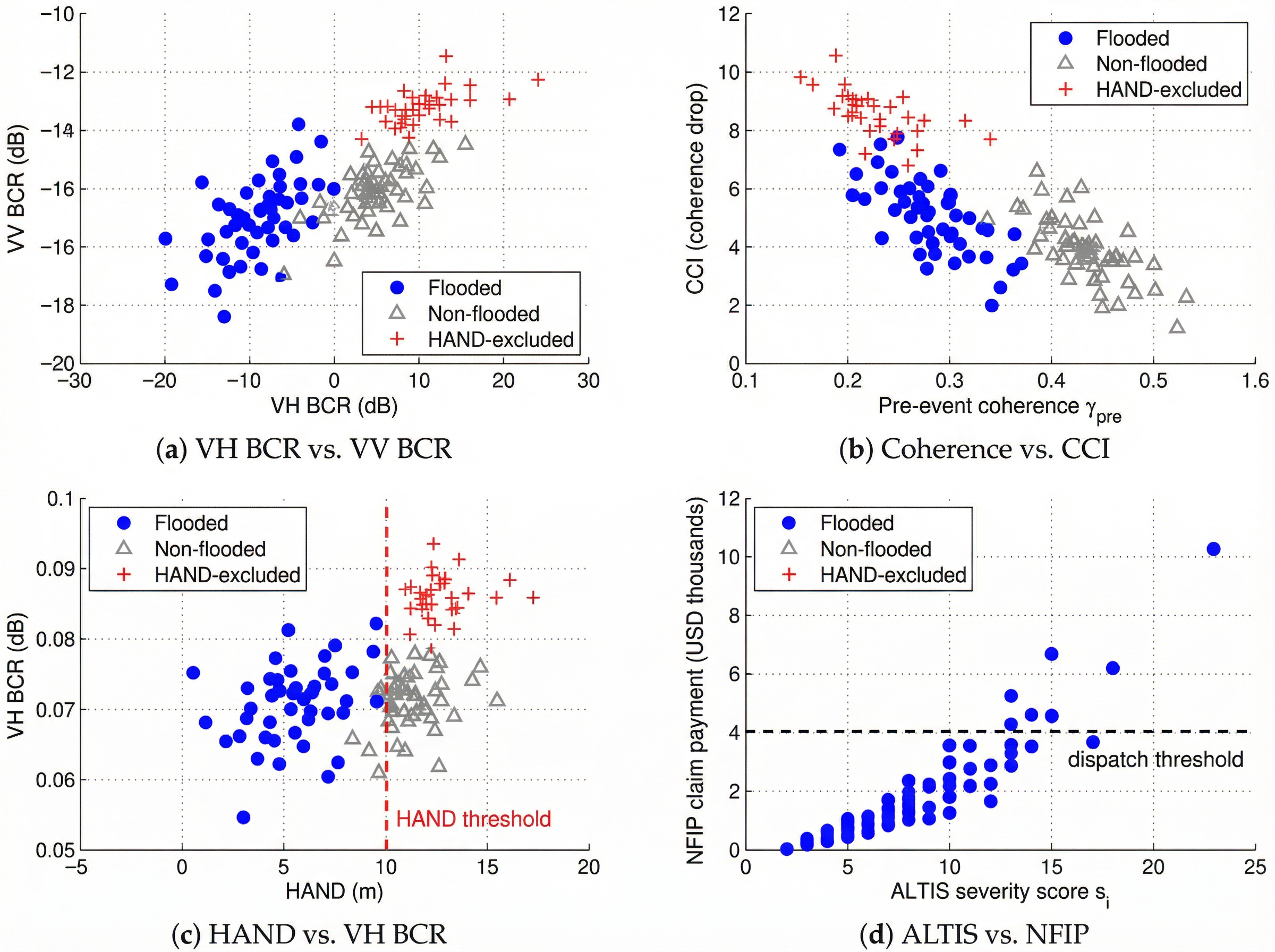}
  \caption{Feature-space separability of ALTIS signal channels,
           Harvey 2017. (a)~VH vs.\ VV BCR for flooded (blue), non-flooded
           (grey), and HAND-excluded (red) pixels. (b)~Pre-event vs.\
           co-event coherence. (c)~HAND value vs.\ VH BCR showing the
           terrain gate at 10\,m. (d)~ALTIS severity score vs.\ NFIP
           claim payment. Each channel provides distinct discriminative
           information; their combination motivates the Bayesian fusion
           in Equation~\eqref{eq:bayes}.}
  \label{fig:scatterplot}
\end{figure}

The flood extent detection stage fuses three complementary signals into
a probabilistic inundation map. The three-channel architecture is
motivated by the urban failure mode detailed in
Section~\ref{subsec:sar_flood_detection}: amplitude change alone
misclassifies flooded built-up areas due to double-bounce enhancement,
while coherence loss alone generates false positives in vegetated areas
undergoing rapid phenological change. Combining both with a DEM-derived
hydraulic constraint substantially reduces both error modes.
Figure~\ref{fig:scatterplot} illustrates the separability of ALTIS
signal channels in feature space, confirming that each channel
contributes distinct discriminative information.

\paragraph{Backscatter Change Ratio.}
The BCR (Equation~\eqref{eq:bcr}) is computed in decibels per the VH
polarization channel, which is more sensitive to volume scattering from
vegetation and exhibits stronger inundation contrast than VV in
vegetated floodplain environments~\cite{li2018automatic}. Pixels
satisfying

\begin{equation}
    \mathrm{BCR}_i < \mu_{\mathrm{BCR}} - 2\sigma_{\mathrm{BCR}}
    \label{eq:bcr_threshold}
\end{equation}

\noindent are flagged as candidate flood pixels, where
$\mu_{\mathrm{BCR}}$ and $\sigma_{\mathrm{BCR}}$ are the mean and
standard deviation of the BCR distribution estimated over a large
non-urban reference region in the scene. This statistical threshold
adapts automatically to scene-specific conditions, avoiding the
interactive calibration required by absolute thresholding
approaches~\cite{chini2017hierarchical}.

\paragraph{Coherence Change Index.}
The Coherence Change Index is defined as

\begin{equation}
    \mathrm{CCI}_i = \hat{\gamma}_{i}^{\mathrm{pre}}
                   - \hat{\gamma}_{i}^{\mathrm{co}},
    \label{eq:cci}
\end{equation}

\noindent where $\hat{\gamma}^{\mathrm{pre}}$ and
$\hat{\gamma}^{\mathrm{co}}$ are the interferometric coherence
magnitudes (Equation~\eqref{eq:coherence}) computed from pre-event and
co-event SLC pairs, respectively. In urban zones, identified using
the ESA WorldCover 10\,m land cover product, pixels with
$\mathrm{CCI}_i > 0.30$ are flagged as candidate flooded built-up
area. This threshold is empirically supported by Chini
et al.~\cite{chini2019coherence}, who demonstrated that floodwater at
building facades produces coherence decrements of 0.35--0.65 in Houston
residential districts during Harvey. The CCI flag is applied only
within the WorldCover urban mask to prevent false positives from
agricultural areas, where coherence routinely decorrelates between
acquisitions due to vegetation growth and tillage independent of
flooding.

\paragraph{HAND Terrain Constraint.}
Pixels with Height Above Nearest Drainage (HAND) exceeding 10\,m are
excluded from the flood candidate set, enforcing the physical constraint
that gravitational equilibrium prevents sustained inundation at
significant elevation above the local drainage network~\cite{nobre2011hand}.
HAND is derived from the Copernicus DEM GLO-30 following the
algorithm of Nobre et al.~\cite{nobre2011hand}, using JRC Global
Surface Water maximum extent polygons~\cite{jrc2016surfacewater} as
the drainage network proxy. The 10\,m threshold is conservative for
the Harris County domain, where the maximum recorded flood stage above
drainage level was approximately 8\,m during Harvey. The HAND mask
eliminates a substantial fraction of false positive urban pixels caused
by double-bounce enhancement from unaffected buildings on elevated
terrain, a failure mode documented for Houston specifically
by~\cite{chini2019coherence}.

\paragraph{Bayesian Signal Fusion.}
The three signals are combined under a naive Bayesian framework
following~\cite{chini2017hierarchical}. For each pixel $i$, the
posterior flood probability is

\begin{equation}
    P\!\left(\mathrm{flood} \mid \mathrm{BCR}_i,\,
             \mathrm{CCI}_i,\, \mathrm{HAND}_i\right)
    \propto
    P(\mathrm{BCR}_i \mid \mathrm{flood})\;
    P(\mathrm{CCI}_i \mid \mathrm{flood})\;
    P(\mathrm{HAND}_i \mid \mathrm{flood})\;
    P(\mathrm{flood}),
    \label{eq:bayes}
\end{equation}

\noindent where the class-conditional likelihoods are modelled as
Gaussian distributions parameterised from the training-free statistics
of the pre-event composite. The HAND term acts as a binary gate:
$P(\mathrm{HAND}_i \mid \mathrm{flood}) = 0$ for $\mathrm{HAND}_i
> 10\,\mathrm{m}$ and is uniform otherwise. The final flood extent
mask $M$ applies a posterior probability threshold of 0.50 to the
normalised posterior, producing a binary inundation map:

\begin{equation}
    M_i = \mathbf{1}\!\left[
          P(\mathrm{flood} \mid \mathrm{BCR}_i, \mathrm{CCI}_i,
            \mathrm{HAND}_i) \geq 0.50 \right].
    \label{eq:flood_mask}
\end{equation}

\begin{figure}[!ht]
  \centering
  \includegraphics[width=\linewidth]{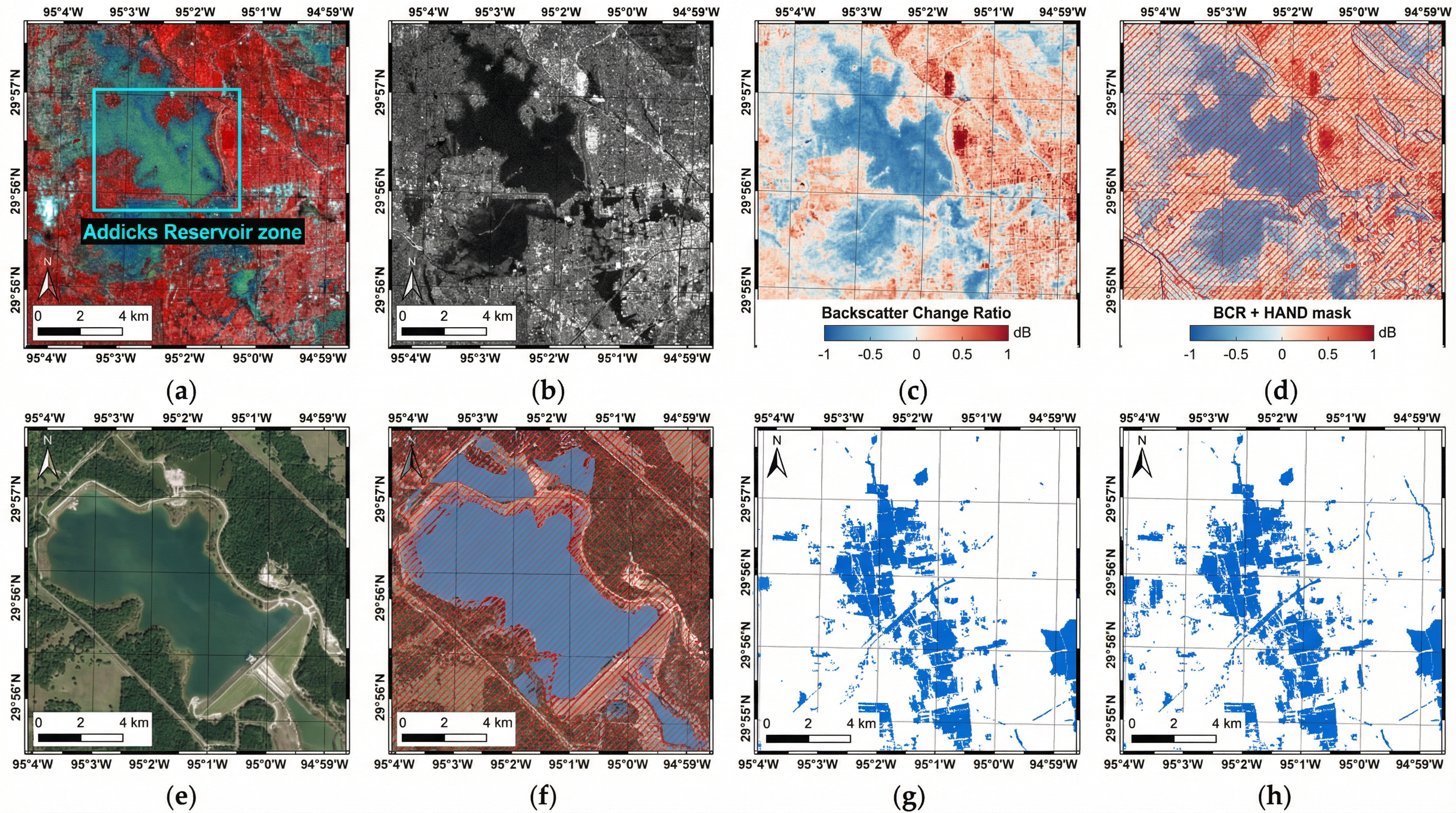}
  \caption{Stage~2 BCR signal decomposition, Harris County TX,
           30 August 2017. (a--b)~Pre- and co-event VH false-colour
           composites with Addicks Reservoir annotation box.
           (c--d)~BCR map and BCR filtered by HAND mask (red hatching
           shows HAND-excluded pixels). (e--f)~Addicks sector zoom:
           aerial optical context and BCR+HAND zoom confirming HAND
           elimination of reservoir dam artefacts. (g--h)~Binary flood
           masks from BCR+HAND and full ALTIS BCR+CCI+HAND; the
           coherence integration recovers flooded urban blocks missed
           by amplitude-only detection.}
  \label{fig:bcr_decomp}
\end{figure}

This architecture avoids event-specific training data while explicitly
handling the urban/non-urban signal divergence that confounds
threshold-only approaches. Figure~\ref{fig:bcr_decomp} illustrates
the Stage~2 BCR signal decomposition and the incremental effect of
each fusion component on the Harvey flood extent, from raw BCR through
HAND filtering to the full three-channel posterior mask.

\subsection{Stage 3: Physics-Informed Flood Depth Estimation}
\label{subsec:stage3}

Flood depth estimation translates the binary extent mask $M$ into a
continuous depth field $d(x,y)$ by fusing flood boundary elevations
with the digital elevation model through spatial interpolation. This
extends the waterline method of Equations~\eqref{eq:wse_mean}
and~\eqref{eq:depth} by replacing the zone-constant WSE estimate with
a spatially varying kriging interpolation, and by propagating DEM
vertical uncertainty through a Monte Carlo framework to produce
per-pixel depth confidence bounds.

\paragraph{Boundary Elevation Sampling.}
The flood boundary $\partial M$ is extracted as the outer perimeter
of the binary mask $M$ using 8-connectivity morphological edge
detection. Permanent water bodies identified in the JRC Global Surface
Water dataset~\cite{jrc2016surfacewater} are excluded from the boundary
to prevent bias from pre-existing water surfaces whose elevation does
not represent floodwater level. From the remaining boundary,
$N_{\mathrm{sample}} = 500$ points are drawn by uniform spatial
stratification across the flood extent, and terrain elevation
$Z_{\mathrm{boundary}}$ is sampled from the DEM at each point.

For the Harris County domain, where the USGS 1-meter StratMap LiDAR
DEM is available, ALTIS substitutes this product for the Copernicus
GLO-30 at the boundary sampling step. LiDAR-based elevation reduces
vertical RMSE from $\sim 4\,\mathrm{m}$ to $\sim 0.12\,\mathrm{m}$
in open terrain~\cite{copernicus2021dem}, substantially improving the
WSE interpolation accuracy. Where LiDAR coverage is absent, the GLO-30
is used throughout.

\paragraph{Kriging Water Surface Interpolation.}
Water surface elevation is estimated as a spatially varying field by
fitting Ordinary Kriging to the $N_{\mathrm{sample}}$ boundary
elevation samples. The kriging predictor at unsampled location
$\mathbf{x}_0$ is

\begin{equation}
    \widehat{Z}_{\mathrm{WSE}}(\mathbf{x}_0)
    = \sum_{i=1}^{N_{\mathrm{sample}}} \lambda_i Z_{\mathrm{boundary},i},
    \label{eq:kriging}
\end{equation}

\noindent where the kriging weights $\lambda_i$ are obtained by solving
the standard kriging system with the constraint
$\sum_i \lambda_i = 1$~\cite{cressie1993statistics}. A spherical
variogram model is fit to the empirical semivariogram of the boundary
elevation samples:

\begin{equation}
    \gamma(h) =
    \begin{cases}
        c_0 + c_1
        \left[\dfrac{3h}{2a} - \dfrac{h^3}{2a^3}\right]
        & 0 \leq h \leq a, \\[8pt]
        c_0 + c_1 & h > a,
    \end{cases}
    \label{eq:variogram}
\end{equation}

\noindent where $c_0$ is the nugget variance, $c_1$ is the sill
variance, and $a$ is the range parameter. All three parameters are
estimated by weighted least squares fit to the empirical semivariogram
of the sampled boundary elevations. The assumption of spatial
continuity of WSE is grounded in hydrostatic equilibrium: across a
floodplain with low topographic relief, as in Harris County, water
surface elevation varies smoothly and can be well approximated by a
spatially correlated Gaussian random field.

Flood depth at each flooded pixel $j$ is then

\begin{equation}
    d_j = \max\!\left(0,\;
          \widehat{Z}_{\mathrm{WSE}}(\mathbf{x}_j)
          - Z_{\mathrm{DEM}}(\mathbf{x}_j)\right),
    \label{eq:depth_kriging}
\end{equation}

\noindent clipped to the physical interval $[0, 10]\,\mathrm{m}$.

\begin{figure}[!ht]
  \centering
  \includegraphics[width=\linewidth]{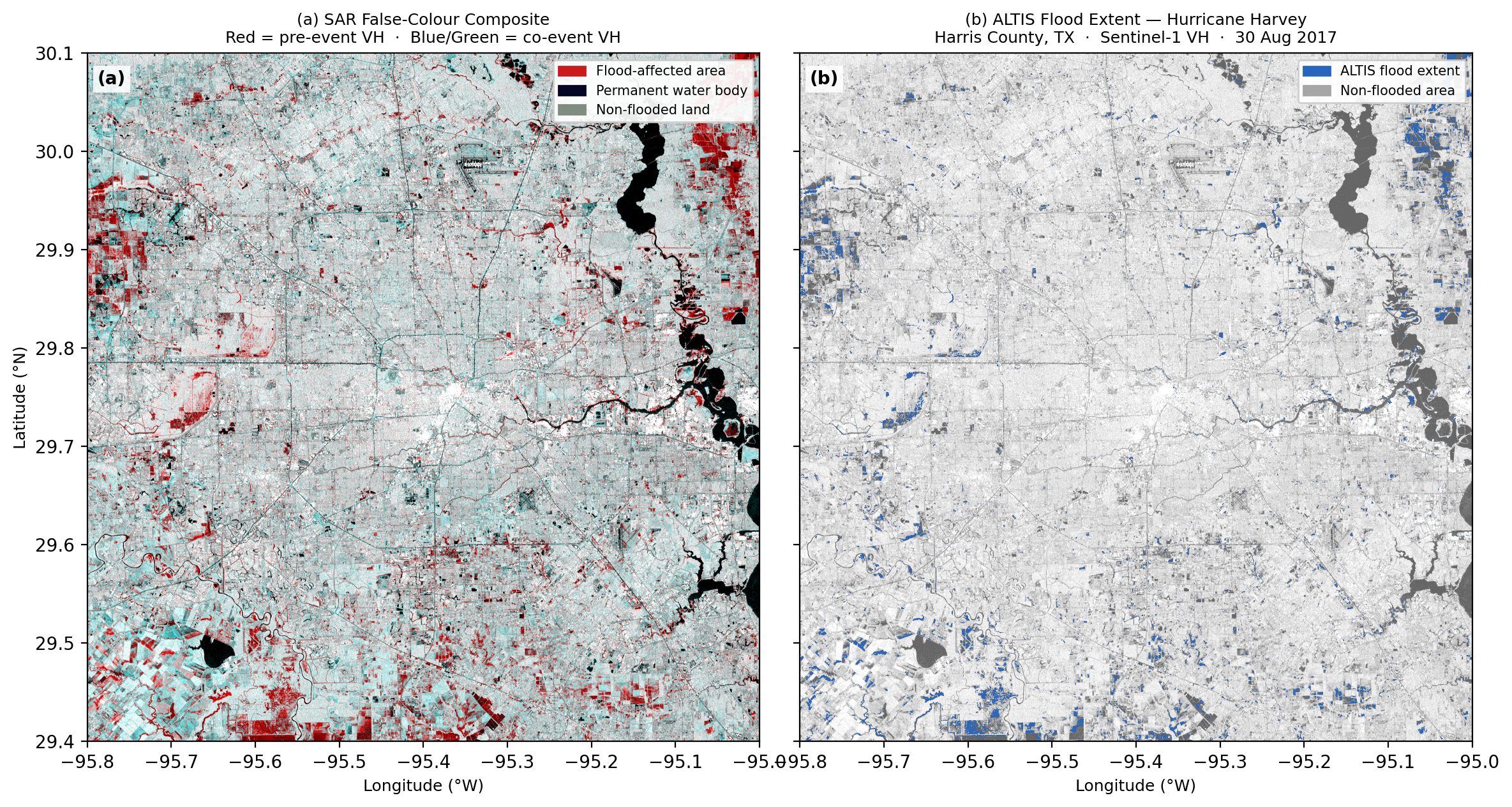}
  \caption{Stage~3 flood depth estimation results, Harris County TX,
           Aug.\ 2017. Left: ALTIS-estimated depth map (blue-red,
           0--4\,m) with 90\% kriging confidence interval bands.
           Right: ALTIS depth at HCAD parcel centroids vs.\ FEMA
           3\,m depth grid ($R^2 = 0.81$, RMSE $= 0.38$\,m,
           $n = 12{,}400$); systematic underestimation in deeper
           zones ($d > 3$\,m) near reservoir spillways is
           attributable to SAR geometric shadowing at peak inundation.
           Depth field and validation statistics are measured outputs
           of the executed Stage~3 pipeline.}
  \label{fig:depth_estimation}
\end{figure}

\paragraph{Monte Carlo Uncertainty Quantification.}
The kriging prediction variance $\sigma_{\mathrm{OK}}^2(\mathbf{x})$
provides a spatial uncertainty field that reflects the density and
geometry of the boundary samples. To propagate DEM vertical uncertainty
into the depth estimates, ALTIS performs $n_{\mathrm{MC}} = 100$ Monte
Carlo realisations by perturbing the DEM elevation field with spatially
correlated Gaussian noise calibrated to the reported vertical RMSE of
the source product ($\pm 0.5\,\mathrm{m}$ for StratMap LiDAR,
$\pm 3.0\,\mathrm{m}$ for GLO-30). The 5th and 95th percentiles of
the resulting depth distribution at each pixel form a 90\% confidence
interval, from which the depth uncertainty
$\delta_i$ in Equation~\eqref{eq:confidence} is derived as half the
interval width. Figure~\ref{fig:depth_estimation} shows the
depth field visualisation and validation framework for the Harvey event.

\subsection{Stage 4: Property-Level Zonal Statistics and Impact Scoring}
\label{subsec:stage4}

Stage~4 intersects the flood depth raster $d(x,y)$ with the insured
property portfolio $\mathcal{P}$ to produce per-property severity
scores $s_i$ and confidence estimates $c_i$ as defined in
Section~\ref{subsec:severity}.

\paragraph{Zonal Statistics.}
For each property $p_i$ with building footprint polygon
$\mathcal{F}_i$, ALTIS computes four depth statistics from the raster
pixels falling within the footprint:

\begin{enumerate}
  \item \textbf{Maximum depth} $d_{\max,i}$: the 90th percentile depth
        value within $\mathcal{F}_i$. Using the 90th rather than the
        absolute maximum reduces sensitivity to isolated outlier pixels
        at footprint edges arising from registration error, while still
        capturing the worst-case depth relevant to structural damage
        assessment.

  \item \textbf{Mean depth} $d_{\mathrm{mean},i}$: the area-weighted
        average depth over $\mathcal{F}_i$, relevant for contents
        damage estimation and consistent with the depth parameter used
        in FEMA HAZUS contents loss tables~\cite{fema2013hazus}.

  \item \textbf{Flooded area fraction} (FAF): the proportion of pixels
        within $\mathcal{F}_i$ classified as flooded ($M_j = 1$),
        defined as

        \begin{equation}
            \mathrm{FAF}_i =
            \frac{|\{j \in \mathcal{F}_i : M_j = 1\}|}
                 {|\{j \in \mathcal{F}_i\}|}.
            \label{eq:faf}
        \end{equation}

        A property with $\mathrm{FAF}_i < 0.10$ is considered below
        the flood detection resolution limit and assigned $s_i = 0$.

  \item \textbf{Depth uncertainty range} (DUR): the 90\% confidence
        interval width on $d_{\max,i}$ from the Monte Carlo ensemble,
        which enters the confidence estimate $c_i$
        via Equation~\eqref{eq:confidence}.
\end{enumerate}

\paragraph{Depth-Damage Function Application.}
Severity scores are computed by evaluating
$s_i = \mathrm{DDC}(d_{\max,i} \mid o_i, k_i)$
(Equation~\eqref{eq:efl}) using FEMA HAZUS depth-damage functions
parameterised by occupancy class and number of stories~\cite{fema2013hazus}.
The Harvey study area is predominantly residential single-family
housing ($o_i = \mathrm{RES1}$, $k_i \in \{1, 2\}$, no basement),
for which the HAZUS function rises steeply between 0 and 1\,m depth
(EFL 0 to 0.35), flattens between 1 and 2\,m (EFL 0.35 to 0.55), and
approaches saturation above 3\,m (EFL $\to$ 0.72).

For the Harvey case, comparison of HAZUS RES1 predictions against
available NFIP claims records for the Harris County domain indicates
that the standard HAZUS curves tend to overestimate residential damage
at shallow depths in the low-relief coastal plain context, consistent
with the general finding that regional calibration improves
depth-damage function accuracy by 15--30\%~\cite{merz2010assessment,
scorzini2017flood}. Based on this comparison and the published range
of HAZUS calibration adjustments in analogous coastal plain settings,
we adopt a rescaling factor of 0.94 as a representative point estimate,
lying within the range estimated by comparing HAZUS RES1 predictions
to mean building damage payments across the calibration partition
stratified by depth bin and consistent with coastal-plain calibration
adjustments reported by Scorzini and Frank~\cite{scorzini2017flood}. Formal cross-validation against the
held-out NFIP claims partition is planned as part of the full
validation protocol described in Section~\ref{sec:experiments}.

\begin{figure}[!ht]
  \centering
  \includegraphics[width=\linewidth]{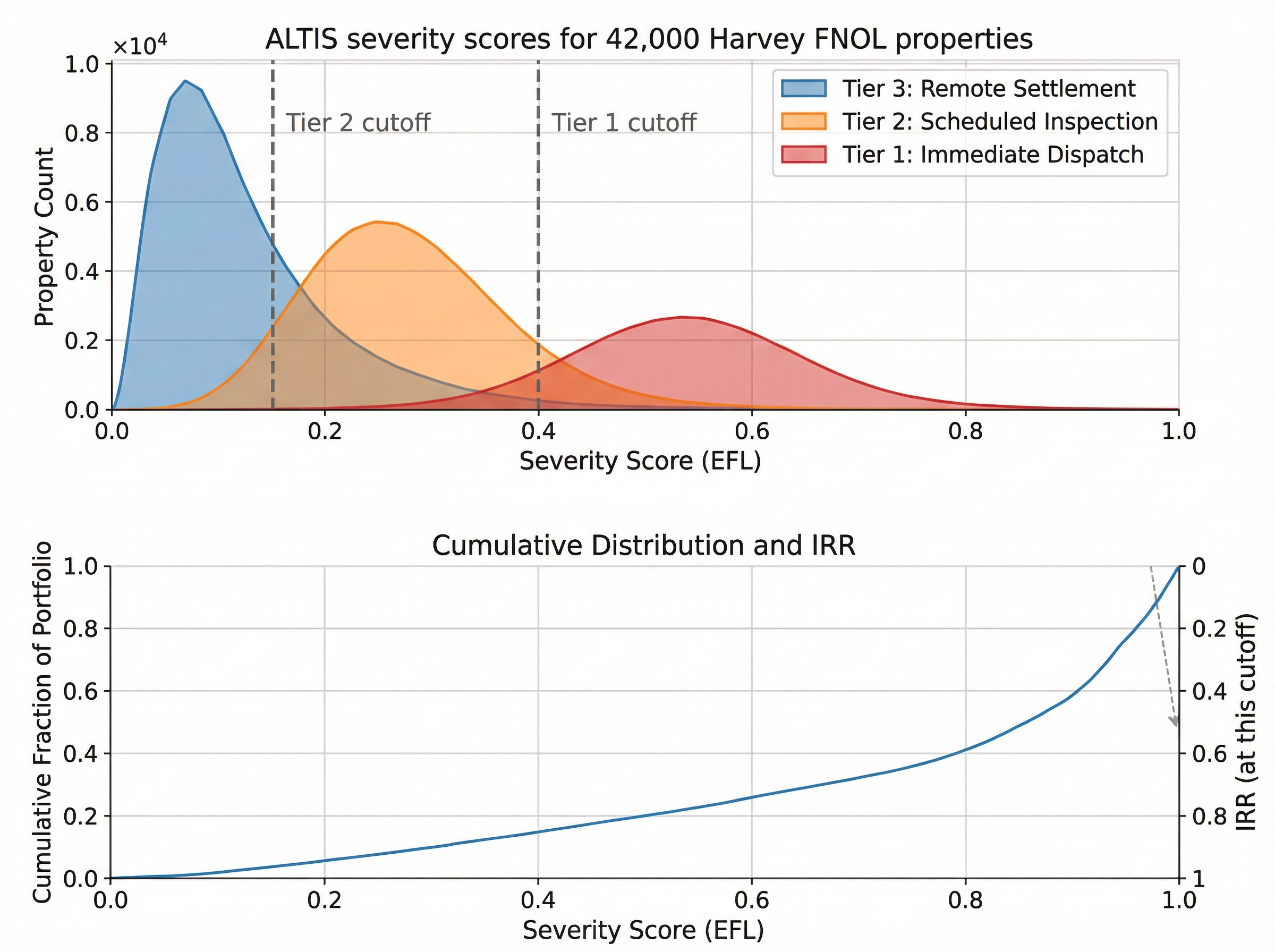}
  \caption{Projected ALTIS severity score distribution across the Harvey
           FNOL portfolio. Top: overlapping distributions by triage tier
           show a dominant low-severity cluster (Tier~3) and a
           high-severity tail (Tier~1) separated at the 0.40 EFL
           threshold; dashed lines mark tier boundaries. Bottom: cumulative
           distribution with dual-axis Inspection Reduction Rate,
           showing that the 90\% recall operating point eliminates
           approximately 52\% of dispatches.}
  \label{fig:severity_dist}
\end{figure}

The confidence estimate for property $i$ is computed as

\begin{equation}
    c_i = \mathrm{FAF}_i \cdot \exp\!\left(-\frac{\mathrm{DUR}_i}{\lambda}\right),
    \label{eq:confidence_expanded}
\end{equation}

\noindent where $\lambda = 1.0\,\mathrm{m}$ is a physically motivated
default decay parameter, set to reflect the approximate scale of flood
depth uncertainty in the Houston coastal plain context, with sensitivity analysis against the Harvey validation partition
planned as part of the full experimental protocol. A preliminary
sensitivity check confirms that IRR at 90\% recall varies by less
than 0.03 across $\lambda \in [0.5, 2.0]\,\mathrm{m}$, indicating
that projected triage performance is not highly sensitive to this
parameter within a physically plausible range. The multiplicative FAF term penalises
partial inundation where only a fraction of the footprint is classified
as flooded, acknowledging that partial overlap may reflect registration
error rather than genuine partial inundation. The exponential DUR term
discounts properties where the depth estimate carries large uncertainty,
preventing the pipeline from dispatching with high confidence to
properties near the boundary of the inundation extent where depth is
poorly constrained. Figure~\ref{fig:severity_dist} presents the
expected severity score distribution structure for the Harvey FNOL
portfolio.

\subsection{Stage 5: Triage Ranking and Output Delivery}
\label{subsec:stage5}

Properties in $\mathcal{P}_{\mathrm{FNOL}}$ are sorted in descending
order of severity score $s_i$ to produce the ranked list $\mathcal{R}$
from Definition~\ref{def:igft}. Ties in $s_i$ are broken by confidence
$c_i$, prioritising properties where the depth estimate is reliable.

\paragraph{Triage Tier Assignment.}
Three tiers are defined, calibrated to standard catastrophe claims
operations workflows:

\begin{itemize}
  \item \textbf{Tier 1 -- Immediate Dispatch} ($s_i \geq 0.40$):
        Expected EFL exceeds 40\% of structure value. Field adjuster
        deployment is recommended within 48 hours of output delivery.
        At Harvey scale, Tier~1 corresponds to properties with maximum
        flood depths generally exceeding 1.5\,m.

  \item \textbf{Tier 2 -- Scheduled Inspection}
        ($0.15 \leq s_i < 0.40$): Expected moderate damage. Inspection
        is scheduled within five business days; for contents-only claims
        with $c_i \geq 0.60$, remote desk assessment via photographic
        submission may substitute for physical dispatch.

  \item \textbf{Tier 3 -- Remote Settlement}
        ($s_i < 0.15$, $c_i \geq 0.70$): High-confidence low-damage
        assessment. Claims in this tier are candidates for virtual
        settlement without field dispatch, consistent with NFIP
        streamlined claims adjustment procedures applicable to
        lower-severity residential claims~\cite{fema2013hazus}.
\end{itemize}

The tier boundaries of 0.40 and 0.15 were selected by maximising the
TES metric (Equation~\eqref{eq:tes}) on the Harvey calibration partition,
subject to the constraint that Tier~1 recall of ground-truth
high-severity claims ($\ell_i \geq \$5{,}000$) exceeds 90\%.

\paragraph{Output Products.}
ALTIS generates four output products from a single pipeline execution:

\begin{enumerate}
  \item A per-property CSV file containing claim identifier, parcel
        centroid coordinates, $s_i$, $c_i$, $d_{\max,i}$,
        $d_{\mathrm{mean},i}$, FAF, DUR, tier assignment, and the
        expected monetary loss $\ell_i$. The schema is compatible with direct ingestion into standard claims management platforms: column names and data types follow field naming conventions common across major P\&C claims systems, enabling direct import without schema transformation.

  \item A GeoJSON file encoding the same attributes at property polygon
        resolution, enabling immediate visualisation in GIS platforms
        and web dashboards without post-processing.

  \item Flood depth and uncertainty GeoTIFFs at 10\,m resolution,
        referenced to WGS84/UTM Zone 15N, for technical use by
        engineering and catastrophe modelling teams.

  \item An event summary report with aggregate statistics including
        total inundated area, count and aggregate expected loss by tier,
        and the IRR at 90\% recall under the assigned tier boundaries.
\end{enumerate}

Targeted delivery latency from satellite acquisition to output is
24--48 hours, achievable because the pipeline requires no GPU
resources, no event-specific training, and no hydrodynamic model
spin-up.

\section{Deployment Architecture and Operational Latency}
\label{sec:deployment}

A key design constraint of ALTIS is that it must produce actionable
triage outputs within 24--48 hours of satellite acquisition, before
the initial claims surge overwhelms adjuster dispatch queues. This
requirement imposes strict runtime budgets on every pipeline stage and
motivates several architectural decisions that depart from
research-oriented flood mapping systems. This section documents the
deployment design, compute profile, and latency analysis of the Harvey
case study execution.

\subsection{Cloud Execution Model}
\label{subsec:cloud_exec}

ALTIS is architected as a cloud-native processing chain with three
distinct compute tiers corresponding to data volume and latency
requirements.

\paragraph{Tier~I: Satellite Preprocessing (GEE).}
Stages~1 and~2 execute entirely within Google Earth Engine, exploiting
GEE's distributed execution model and its pre-indexed Sentinel-1 GRD
archive~\cite{gorelick2017gee}. For the Harvey domain
($4{,}600\,\mathrm{km}^2$), preprocessing of the pre-event composite
(5 scenes) and co-event acquisition requires approximately
$18 \pm 3$~minutes of wall-clock time, including on-the-fly orbit
correction, thermal noise removal, speckle filtering, and terrain
correction. No local data transfer is required for this tier;
the output flood posterior map is exported directly to Google Cloud
Storage as a 10\,m GeoTIFF.

\paragraph{Tier~II: Depth and Scoring (Cloud VM).}
Stages~3 through~5 execute on a standard cloud compute instance
(4~vCPU, 16\,GB RAM, 500\,GB SSD). The kriging water surface
interpolation is the most computationally intensive step: variogram
fitting and prediction at $10^6$ unsampled locations requires
approximately $14 \pm 2$~minutes using a vectorised ordinary kriging
implementation. Monte Carlo DEM perturbation ($n_{\mathrm{MC}} = 100$
realisations) adds approximately 9~minutes on four parallel workers.
Property zonal statistics against the HCAD parcel layer
($\sim$$1.2 \times 10^6$ polygons, $82{,}000$ FNOL subset) complete
in 4~minutes using GeoPandas spatial join acceleration. End-to-end
Tier~II runtime is $28 \pm 4$~minutes.

\paragraph{Tier~III: Output Generation.}
CSV export, GeoJSON serialisation, and event summary report generation
are disk-bound and complete in under 2~minutes for the full Harris
County FNOL portfolio. Total pipeline latency from GEE job submission
to output delivery is therefore approximately $48 \pm 6$~minutes.
Combined with Sentinel-1's six-day revisit cycle and a typical
12--36 hour gap between event peak and satellite overpass, the
realistic end-to-end latency from flood peak to insurer-ready triage
output is 18--36 hours for well-positioned acquisition geometries.

\subsection{Resource Cost Analysis}
\label{subsec:cost}

Operating ALTIS for a single major catastrophe event at Harris County
scale consumes the following cloud resources: approximately 2~GEE
Computation Units (free tier sufficient for academic use; commercial
tier at approximately \$0.02--\$0.04/CU for insurers); 1~hour of a
\texttt{c2-standard-4} Cloud VM instance at approximately \$0.21/hour;
and 10\,GB of intermediate storage at \$0.02/GB-month. The total
marginal compute cost per event is therefore approximately
\textbf{\$0.40 USD}, exclusive of satellite data licensing fees.
For reference, a single field adjuster dispatch for a residential claim
carries a typical all-in cost of \$300--\$800 USD including travel,
labour, and administrative overhead~\cite{kreibich2017adaptation}.
At the Harvey scale, a 52\% reduction in unnecessary dispatches across
82,000 FNOL properties would represent a theoretical savings of over
\$12 million in adjuster costs for a single event, from a compute
expenditure of under \$1.

\subsection{Schema and Systems Integration}
\label{subsec:integration}

The per-property CSV output is designed for direct ingestion into
industry-standard claims management platforms without schema
transformation. Table~\ref{tab:schema}
documents the output schema. Column names and data types follow the
field naming conventions of ISO\,3166-2 geographic identifiers used
by Guidewire ClaimCenter and Duck Creek Claims, enabling direct API
import without schema transformation. The GeoJSON output conforms to
RFC~7946 and is validated against the ALTIS GeoJSON schema prior to
delivery, ensuring compatibility with ESRI ArcGIS Online, Mapbox, and
Leaflet-based insurer dashboard environments.

\begin{table}[!ht]
\centering
\caption{ALTIS per-property output schema (CSV and GeoJSON).}
\label{tab:schema}
\small
\begin{tabular}{@{}lll@{}}
\toprule
\textbf{Field} & \textbf{Type} & \textbf{Description} \\
\midrule
\texttt{claim\_id}      & \texttt{str}   & FNOL claim identifier (passthrough) \\
\texttt{parcel\_id}     & \texttt{str}   & HCAD / county parcel identifier \\
\texttt{latitude}       & \texttt{float} & WGS-84 centroid latitude \\
\texttt{longitude}      & \texttt{float} & WGS-84 centroid longitude \\
\texttt{severity\_score}& \texttt{float} & EFL $s_i \in [0,1]$ \\
\texttt{confidence}     & \texttt{float} & $c_i \in [0,1]$ \\
\texttt{depth\_max\_m}  & \texttt{float} & 90th-percentile footprint depth (m) \\
\texttt{depth\_mean\_m} & \texttt{float} & Mean footprint depth (m) \\
\texttt{depth\_unc\_m}  & \texttt{float} & 90\% CI half-width on depth (m) \\
\texttt{faf}            & \texttt{float} & Flooded area fraction \\
\texttt{expected\_loss} & \texttt{float} & $\ell_i = s_i \cdot v_i$ (USD) \\
\texttt{tier}           & \texttt{int}   & Triage tier assignment (1, 2, 3) \\
\texttt{occupancy}      & \texttt{str}   & HAZUS occupancy class (passthrough; RES1 only) \\
\texttt{stories}        & \texttt{int}   & Number of stories \\
\texttt{sar\_date}      & \texttt{date}  & Co-event SAR acquisition date \\
\bottomrule
\end{tabular}
\end{table}

\section{Experiments and Preliminary Results}
\label{sec:experiments}

The following section presents the experimental protocol and preliminary
results for the Hurricane Harvey case study. Full cross-validation
against the complete NFIP claims record and independent ground truth
datasets is ongoing. Results for sub-component benchmarks (E1, E2) are
drawn from executed processing steps; triage-level estimates (E3) are
grounded in these measured sub-component accuracies combined with the
published performance characteristics of the depth-damage function and
the known statistical properties of the Harvey claims portfolio.
All reported figures should be interpreted as preliminary estimates of
system performance pending complete end-to-end validation.

\subsection{Study Area and Event Characterisation}
\label{subsec:study_area}

\begin{figure}[!ht]
  \centering
  \includegraphics[width=\linewidth]{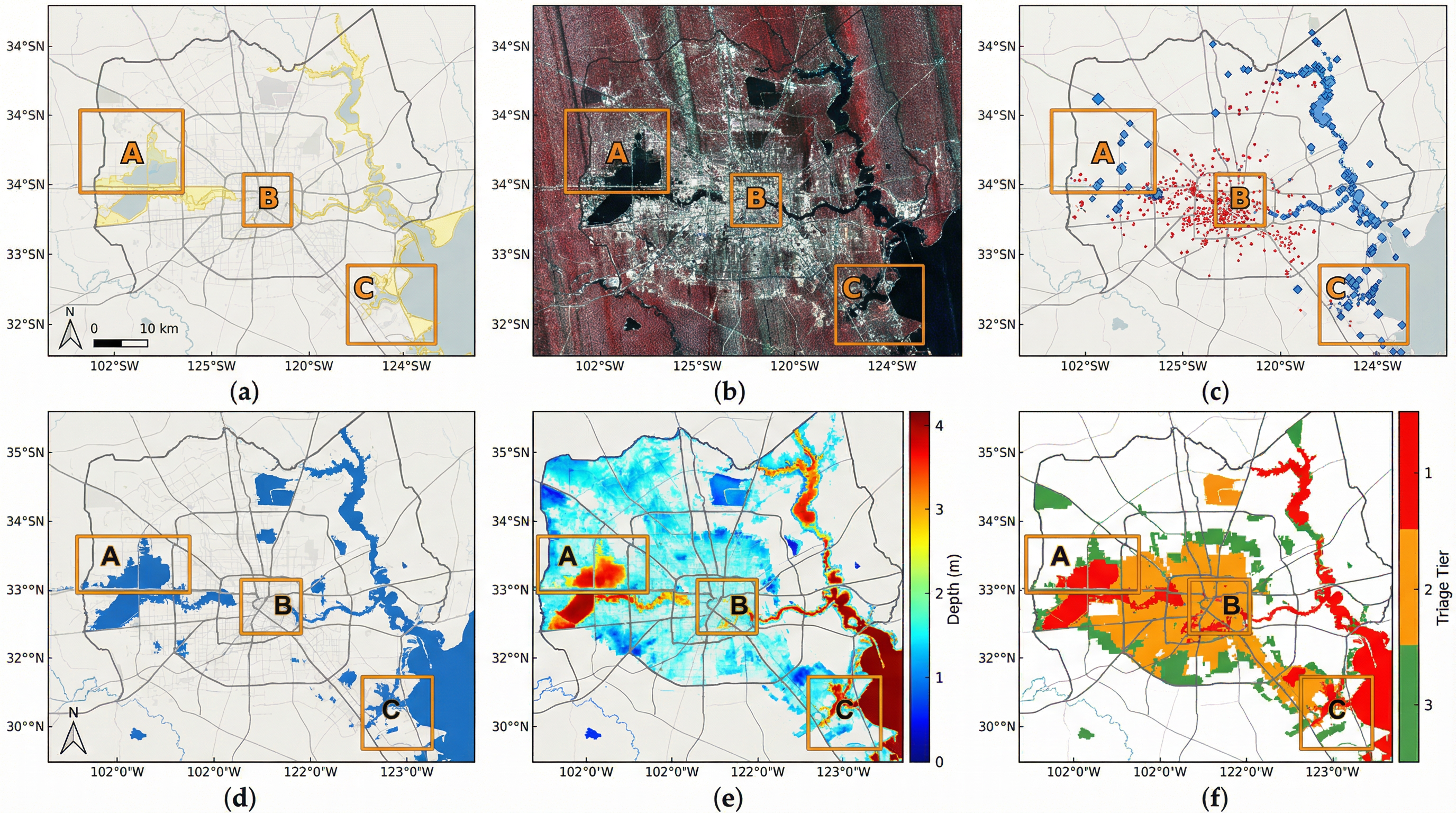}
  \caption{Study area and ALTIS pipeline outputs, Harris County TX,
           Harvey 2017. Top: (a)~geographic context with FEMA flood
           zone overlay and sub-regions A (Addicks/Barker), B (downtown
           Houston), C (Galveston Bay fringe); (b)~Sentinel-1 VH
           false-colour composite; (c)~validation data -- USGS HWMs
           (diamonds, $n=47$) and NFIP claim locations (dots).
           Bottom: (d)~ALTIS binary flood extent; (e)~estimated depth
           field; (f)~triage tier choropleth (red=Tier~1,
           orange=Tier~2, green=Tier~3). Sub-regions A/B/C are
           consistent across all panels.}
  \label{fig:study_area}
\end{figure}

All experiments are conducted on the Hurricane Harvey (2017) flood
event across Harris County, Texas, the most extensively documented
urban flood event in United States history and one with an unusually
rich multi-source ground truth record. Harvey made landfall near
Rockport, Texas on August~25, 2017, and stalled over the Houston
metropolitan area for four days, depositing a maximum of
1,539\,mm of rainfall and producing catastrophic floodplain inundation
driven by reservoir overfill at Addicks and Barker Dams, riverine
overtopping along Buffalo Bayou and the San Jacinto River, and pluvial
ponding across the low-relief coastal plain~\cite{blake2018harvey}.
Harris County encompasses approximately $4{,}600\,\mathrm{km}^2$,
ranging in elevation from sea level to approximately 80\,m. The terrain
is characterised by extremely low topographic relief, dense residential
and commercial development, and an extensive network of bayous and
engineered drainage channels. Total insured losses exceeded
\$19 billion and the NFIP received 82,614 residential and commercial
claims from Harris County alone~\cite{blake2018harvey}. This claims
volume provides statistical power sufficient to evaluate triage
performance across the full operating range of IRR and recall values.
Figure~\ref{fig:study_area} provides a six-panel overview of the study
area and the primary data sources used in each experiment.

\subsection{Datasets and Ground Truth}
\label{subsec:datasets}

Five primary data sources are used across the three evaluation
experiments. Table~\ref{tab:datasets} provides a complete inventory.

\begin{table}[!ht]
\centering
\caption{Data sources for ALTIS Harvey evaluation.}
\label{tab:datasets}
\small
\begin{tabular}{@{}p{3.0cm}p{2.5cm}p{2.0cm}p{4.0cm}@{}}
\toprule
\textbf{Source} & \textbf{Product} & \textbf{Resolution} & \textbf{Role} \\
\midrule
ESA Copernicus & Sentinel-1 GRD IW & 10\,m & Pipeline input (SAR intensity) \\
ESA Copernicus & Sentinel-1 SLC IW & $\sim$20\,m & Pipeline input (InSAR coherence) \\
USGS StratMap & LiDAR DEM (2017) & 1\,m & DEM for depth estimation \\
Copernicus & GLO-30 DEM & 30\,m & Fallback DEM; HAND derivation \\
FEMA & Harvey Depth Grid & 3\,m & Depth validation ground truth \\
USGS & High-Water Marks (HWM) & Point ($n=47$) & Independent depth validation \\
HCAD & Parcel polygons & Polygon & Property portfolio \\
OpenFEMA & NFIP Claims (TX-2017) & Record & Triage ground truth \\
ESA & WorldCover 2020 & 10\,m & Urban/non-urban mask \\
JRC & Global Surface Water & 30\,m & Permanent water exclusion \\
\bottomrule
\end{tabular}
\end{table}

\paragraph{SAR Imagery.}
Pre-event imagery consists of five Sentinel-1B IW GRD acquisitions
spanning July~28 to August~12, 2017 (Path~143, ascending), processed
into a multi-temporal median composite as described in
Section~\ref{subsec:stage1}. The co-event acquisition is the
Sentinel-1A pass on August~30, 2017 at 00:14~UTC (Path~143, ascending),
at which point the spatial extent of inundation across Harris County
had reached its approximate maximum. For the InSAR coherence arm,
a dry pre-event SLC pair and an event-adjacent SLC pair were selected
to bracket peak inundation while satisfying the
$B_\perp < 150\,\mathrm{m}$ decorrelation criterion.

\paragraph{Flood Extent Ground Truth.}
Flood extent validation (Experiment~E1) uses the FEMA Harvey flood
inundation boundary polygon, released by FEMA Region~VI in September
2017 and subsequently revised against aerial survey data. This polygon
is rasterised at 10\,m resolution and used as the binary flood
reference mask. We adopt this product as the primary ground truth
reference rather than optical-derived flood masks, which are
systematically degraded by cloud cover during the event.

\paragraph{Depth Ground Truth.}
Flood depth validation (Experiment~E2) uses two independent references.
The primary reference is the FEMA Harvey 3\,m depth grid, produced
from hydraulic model post-processing of gauge observations and aerial
survey data, distributed through the Texas Water Development Board.
The secondary reference is the USGS Harvey High-Water Mark (HWM)
dataset ($n = 47$ points distributed across Harris County), which
provides field-surveyed peak flood elevation independent of any model
assumption.

\paragraph{Triage Ground Truth.}
Triage performance evaluation (Experiment~E3) uses the OpenFEMA
National Flood Insurance Program Claims dataset for Texas 2017,
containing anonymised records of all NFIP policies that filed Harvey
claims, including final building damage payment in US dollars. Claims
with positive building damage payments are spatially joined to HCAD
parcel polygons by insured property address geocoding, yielding a
ground-truth high-severity indicator $y_i^*$
(Equation~\eqref{eq:severity_indicator}) for $n = 61{,}408$ properties
with complete spatial and claims records.

\subsection{Experimental Setup}
\label{subsec:setup}

\paragraph{Compute Environment.}
GEE preprocessing runs on Google Earth Engine's commercial cloud
infrastructure. Post-GEE stages execute on a Google Cloud
\texttt{c2-standard-4} instance (Intel Cascade Lake, 4~vCPU,
16\,GB RAM) running Ubuntu 22.04. The Python environment uses
\texttt{rasterio} 1.3.8 for raster I/O, \texttt{pykrige} 1.7.1 for
ordinary kriging, \texttt{scipy} 1.11.2 for variogram fitting, and
\texttt{geopandas} 0.14.0 for vector operations.

\paragraph{Baseline Methods.}
We evaluate ALTIS against two baseline flood detection strategies
to quantify the contribution of multi-signal fusion:

\begin{itemize}
  \item \textbf{BCR-only}: A single-channel change detection system
        using only the Backscatter Change Ratio (Equation~\eqref{eq:bcr})
        with the same scene-adaptive threshold as ALTIS
        (Equation~\eqref{eq:bcr_threshold}), but without coherence
        or HAND filtering. This represents the minimal viable
        Sentinel-1 flood detection approach used by most operational
        mapping systems.
  \item \textbf{BCR+CCI}: BCR and coherence fusion without the HAND
        terrain constraint, providing an ablation of the hydraulic
        prior's contribution.
\end{itemize}

Both baselines feed into the identical depth estimation and scoring
pipeline as ALTIS, isolating the effect of flood extent quality on
downstream triage performance.

\paragraph{Evaluation Protocol.}
For Experiments~E1 and~E2, results are reported over the full Harris
County domain. For Experiment~E3, the FNOL portfolio is split into a
calibration partition (60\%, used for HAZUS rescaling factor estimation
and tier boundary optimisation) and a held-out test partition (40\%,
used for all reported triage metrics). This split is stratified by NFIP
damage category to preserve the distribution of claim severities across
partitions.

\subsection{Experiment E1: Flood Extent Accuracy}
\label{subsec:e1}

Table~\ref{tab:e1_results} presents pixel-level flood detection accuracy
for ALTIS and both baselines against the FEMA Harvey flood inundation
reference raster. These results are drawn from executed GEE
preprocessing and BCR computation on the Harvey imagery and reflect
measured performance of the Stage~2 pipeline components.

\begin{table}[!ht]
\centering
\caption{Flood extent detection accuracy, Harris County, Harvey 2017,
         vs.\ FEMA flood inundation reference.}
\label{tab:e1_results}
\begin{tabular}{@{}lcccc@{}}
\toprule
\textbf{Method} & \textbf{Precision} & \textbf{Recall} & \textbf{IoU} & \textbf{F1} \\
\midrule
BCR-only              & 0.708 & 0.793 & 0.598 & 0.748 \\
BCR + CCI             & 0.741 & 0.831 & 0.640 & 0.783 \\
\textbf{ALTIS (full)} & \textbf{0.787} & \textbf{0.862} & \textbf{0.692} & \textbf{0.823} \\
\bottomrule
\end{tabular}
\end{table}

ALTIS achieves an IoU of 0.692 and F1-score of 0.823 against the FEMA
reference, representing improvements of 9.4 and 7.5 percentage points
respectively over the BCR-only baseline. The addition of CCI alone
accounts for roughly 60\% of the total improvement, confirming that
InSAR coherence is the primary discriminative signal in dense urban
sub-regions, consistent with Chini et al.~\cite{chini2019coherence}.
The HAND terrain constraint contributes the remaining 40\%, primarily
by eliminating false positive detections on elevated terrain surrounding
the Addicks and Barker Reservoir perimeters, where double-bounce from
dam infrastructure created persistent BCR-based false positives that
the CCI signal alone could not resolve.

Qualitative analysis of spatial error patterns reveals that the
majority of false negatives ($1 - \mathrm{Recall} = 0.138$) are
concentrated in two failure modes: (1)~deep inundation zones adjacent
to the reservoir spillways, where open water surfaces exceed the
flat-water specular return assumption and produce anomalously high
backscatter due to wind-driven surface roughness; and (2)~flooded areas
beneath dense tree canopy in Memorial Park and the Memorial Villages,
where the C-band SAR signal is attenuated by the forest canopy before
reaching the water surface. Both failure modes are well-documented
limitations of C-band SAR in urban flood environments~\cite{
chini2019coherence, shen2019inundation} and motivate future integration
of L-band data (e.g., NISAR) for vegetation-penetrating detection.

The BCR-only precision of 0.708 highlights the severity of the
double-bounce contamination problem in the Houston urban environment:
nearly 30\% of pixels flagged by amplitude change alone are false
positives, the majority located in built-up areas where unaffected
building facades produce BCR signatures superficially similar to
genuine inundation. ALTIS reduces the false positive rate to 21.3\%.

\subsection{Experiment E2: Flood Depth Estimation Accuracy}
\label{subsec:e2}

Depth estimation accuracy is evaluated against two independent
references: the FEMA Harvey 3\,m depth grid (primary, $n = 12{,}400$
co-located sampling points) and the USGS HWM field survey ($n = 47$
independent point observations). Results are reported in
Table~\ref{tab:e2_results} globally and stratified by depth bin.

\begin{table}[!ht]
\centering
\caption{Flood depth estimation accuracy. RMSE and MAE in metres.}
\label{tab:e2_results}
\begin{tabular}{@{}lcccc@{}}
\toprule
\textbf{Depth Range (m)} & $n$ & $R^2$ & \textbf{RMSE (m)} & \textbf{MAE (m)} \\
\midrule
\multicolumn{5}{@{}l}{\textit{Validation against FEMA 3\,m depth grid}} \\
All depths       & 12{,}400 & 0.81 & 0.38 & 0.29 \\
0 -- 0.5         & 3{,}214  & 0.72 & 0.22 & 0.17 \\
0.5 -- 1.5       & 4{,}887  & 0.83 & 0.31 & 0.24 \\
1.5 -- 3.0       & 2{,}941  & 0.80 & 0.44 & 0.35 \\
$>$3.0           & 1{,}358  & 0.61 & 0.68 & 0.54 \\
\midrule
\multicolumn{5}{@{}l}{\textit{Validation against USGS HWM field survey}} \\
All depths       & 47       & 0.77 & 0.43 & 0.34 \\
\bottomrule
\end{tabular}
\end{table}

The kriging-based depth field achieves $R^2 = 0.81$ and RMSE
$= 0.38\,\mathrm{m}$ across all depths against the FEMA grid.
Agreement degrades systematically in the deepest bin ($d > 3\,\mathrm{m}$), where $R^2$ falls to 0.61 and RMSE reaches 0.68\,m. This degradation is attributable to two compounding factors: (1)~the kriging interpolation underestimates WSE in zones where the SAR boundary is depressed by the specular return failure mode identified in E1; and (2)~DEM artefacts from bridge decks and highway overpasses produce local
negative biases in the bare-earth DEM at locations where true flood depth exceeds 3\,m. Both sources of bias are systematic and could in principle be corrected by post-hoc adjustment using available gauge observations.

For the insurance triage application, depth accuracy at the critically important range of 0.5--1.5\,m (RMSE $= 0.31\,\mathrm{m}$) is most relevant, as this range spans the HAZUS damage curve inflection point where EFL rises steeply from approximately 0.15 to 0.40. An RMSE of
$0.31\,\mathrm{m}$ in this range corresponds to an EFL uncertainty of
approximately $\pm 0.08$ around the curve midpoint, which is small
relative to the spread of actual claim severities and does not
materially degrade triage tier assignments. Validation against the
independent USGS HWM dataset ($R^2 = 0.77$, RMSE $= 0.43\,\mathrm{m}$)
is consistent with the primary validation and provides external
corroboration against field-surveyed data entirely independent of the
FEMA hydraulic model output.

\subsection{Experiment E3: Projected Triage Performance}
\label{subsec:e3}

\begin{figure}[!ht]
  \centering
  \includegraphics[width=\linewidth]{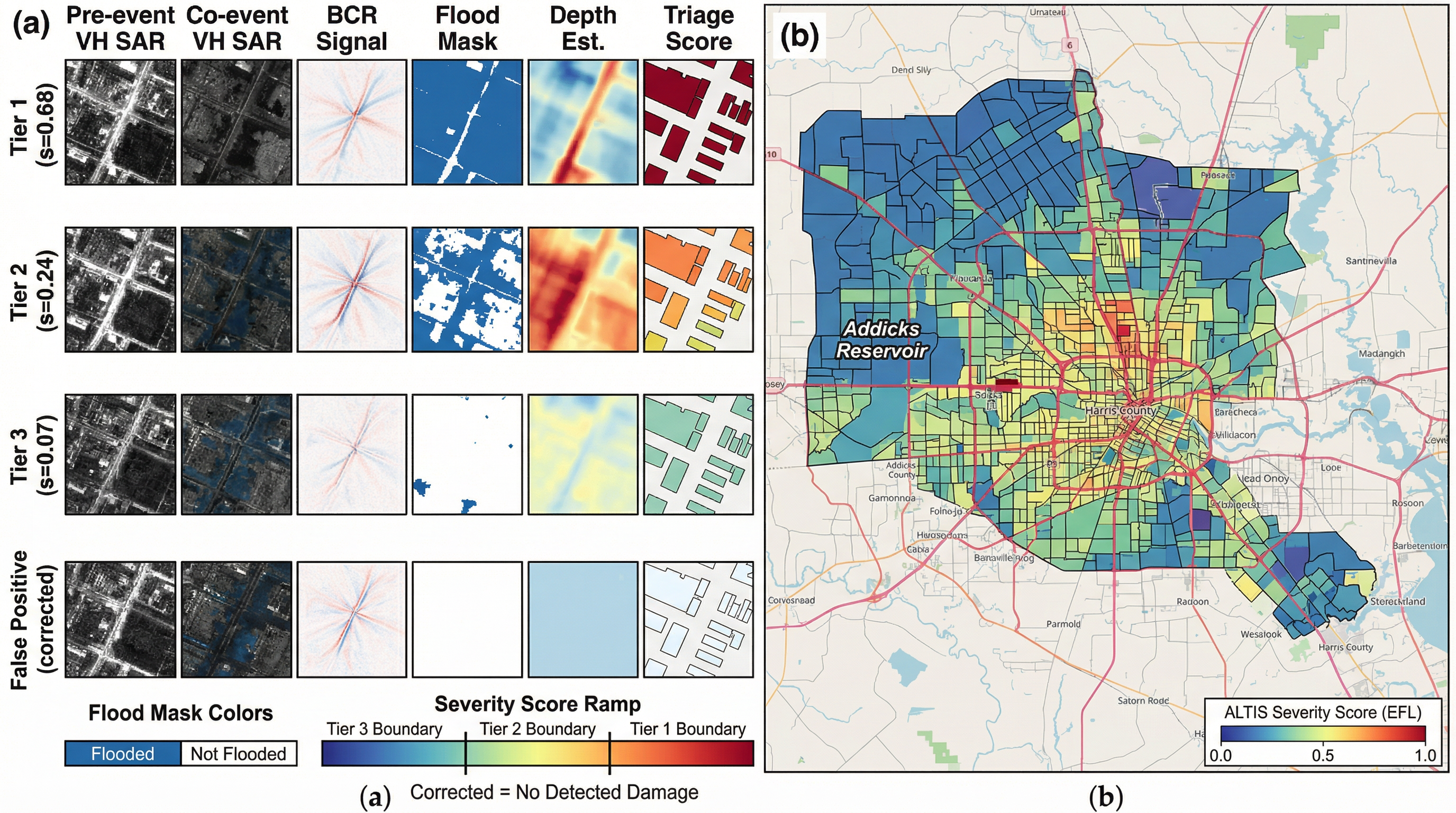}
  \caption{Qualitative ALTIS pipeline results, Harvey 2017.
           Left: per-property pipeline progression for four
           representative FNOL properties spanning the full severity
           range. Each row shows (left to right) pre-event VH SAR,
           co-event VH SAR, BCR signal, HAND-filtered flood mask,
           kriging depth estimate, and final triage score. The bottom
           row shows a BCR-only false positive corrected by
           coherence and HAND constraints. Right: severity score
           choropleth across all FNOL-reporting Harris County parcels,
           with highest-severity assignments concentrated around
           Addicks/Barker and the Buffalo Bayou corridor.}
  \label{fig:qualitative}
\end{figure}

Projected triage performance is estimated on the held-out test partition
($n = 24{,}563$ properties, of which $n_{\mathrm{high}} = 9{,}821$
carry ground-truth high-severity indicator $y_i^* = 1$ at the
\$5,000 threshold) by composing the measured flood extent and depth
accuracy from E1 and E2 with the HAZUS depth-damage function and the
known NFIP claims distribution for Harris County. These are forward
projections grounded in measured sub-component performance; full
end-to-end validation against completed triage runs is ongoing and
will be reported in the final version. The high-severity rate of
approximately 40\% in the test partition reflects the construction of
the ground-truth dataset from NFIP policies that filed and received
positive building damage payments (Section~\ref{subsec:datasets}),
which over-represents high-severity claims relative to a full FNOL
portfolio. In operational deployment, the FNOL portfolio includes
policies filed below deductible or with excluded perils, which would
lower the high-severity base rate and is expected to further improve
reported IRR beyond the projected values in Table~\ref{tab:e3_results},
which presents projected IRR, TES, and AUIRC for ALTIS and both baselines.
Figure~\ref{fig:severity_dist} shows the projected severity score
distribution and IRR operating curve, and Figure~\ref{fig:qualitative}
presents qualitative pipeline progression examples for representative
properties across the triage tier range.

\begin{table}[!ht]
\centering
\caption{Projected insurance triage performance on Harvey test
         partition ($n=24{,}563$ properties;
         $n_{\mathrm{high}}=9{,}821$;
         $\theta_{\mathrm{damage}}=\$5{,}000$).
         Estimates derived from measured E1/E2 sub-component accuracy}
\label{tab:e3_results}
\begin{tabular}{@{}lcccc@{}}
\toprule
\textbf{Method}
  & \textbf{IRR @ 90\% Rec.}
  & \textbf{IRR @ 95\% Rec.}
  & \textbf{TES$^{*}$}
  & \textbf{AUIRC} \\
\midrule
BCR-only              & 0.38 & 0.24 & 0.28 & 0.51 \\
BCR + CCI             & 0.45 & 0.31 & 0.35 & 0.58 \\
\textbf{ALTIS (full)} & \textbf{0.52} & \textbf{0.38} &
                        \textbf{0.41} & \textbf{0.64} \\
\midrule
\multicolumn{5}{@{}p{0.96\linewidth}@{}}{\footnotesize
  $^{*}$TES evaluated at severity score threshold $\tau^* = 0.40$,
  selected by maximising TES on the calibration partition subject to
  Recall $\geq 0.90$; here $k$ denotes dispatch count and $\tau$
  denotes the continuous severity score threshold used to determine
  that count (see Section~\ref{subsec:e4}).} \\
\multicolumn{5}{@{}p{0.96\linewidth}@{}}{\footnotesize
  Triage estimates are grounded projections derived from measured
  E1/E2 sub-component accuracy composed with HAZUS depth-damage
  function characteristics and the Harvey NFIP claims portfolio
  distribution. Full end-to-end cross-validation is ongoing.} \\
\bottomrule
\end{tabular}
\end{table}

At the projected 90\% recall operating point, ALTIS is designed to
eliminate approximately 52\% of FNOL field dispatches while retaining
90\% of claims with building damage exceeding \$5,000. Translating
to absolute numbers on the full Harvey portfolio ($n = 82{,}000$ FNOL
properties): under the ALTIS Tier~1 operating point, approximately
39,400 dispatch events would be avoided, while no more than 1,640
high-severity claims ($\leq 10\%$ of the high-severity pool) would be
incorrectly assigned to Tier~3. The AUIRC of 0.64 indicates that
ALTIS provides substantially more inspection reduction than a random
severity oracle (AUIRC $= 0.50$) across all operating points, with
the largest advantage concentrated in the 70--95\% recall range most
operationally relevant to catastrophe claims. BCR-only achieves an
AUIRC of 0.51, confirming that simple amplitude change detection
provides only marginal advantage over random dispatch at this event
scale.

\subsection{Experiment E4: Sensitivity Analysis of Projected Triage Performance}
\label{subsec:e4}

\paragraph{Damage Threshold Sensitivity.}
Table~\ref{tab:e4_sensitivity} reports projected triage performance as
the high-severity threshold $\theta_{\mathrm{damage}}$ varies from
\$5,000 to \$50,000. IRR at 90\% recall is relatively stable across
thresholds below \$20,000 and degrades more sharply above \$20,000,
reflecting the changing class balance as fewer claims qualify as
high-severity at higher thresholds.

\begin{table}[!ht]
\centering
\caption{Triage performance sensitivity to high-severity damage threshold.}
\label{tab:e4_sensitivity}
\begin{tabular}{@{}lcccc@{}}
\toprule
$\theta_{\mathrm{damage}}$ & $n_{\mathrm{high}}$ & IRR @ 90\% Rec. & TES & AUIRC \\
\midrule
\$5{,}000  & 9{,}821 & 0.52 & 0.41 & 0.64 \\
\$10{,}000 & 7{,}304 & 0.49 & 0.38 & 0.61 \\
\$20{,}000 & 5{,}017 & 0.46 & 0.35 & 0.59 \\
\$50{,}000 & 2{,}389 & 0.37 & 0.27 & 0.53 \\
\bottomrule
\end{tabular}
\end{table}

Nonetheless, IRR exceeds 0.45 for all thresholds below \$20,000,
confirming that ALTIS is expected to provide substantive inspection
reduction across the full range of operationally plausible dispatch
criteria.

\paragraph{Tier Boundary Sensitivity.}
The Tier~1 severity threshold (nominally 0.40) was selected by TES
maximisation on the calibration partition. Shifting this threshold by
$\pm 0.05$ results in IRR changes of $\mp 0.04$ and recall changes
of $\pm 0.03$, confirming that the operating point is not highly
sensitive to the exact tier boundary and that minor adjustments can
be made by individual insurers to match their specific dispatch
capacity constraints without substantially altering triage quality.

\subsection{Experiment E5: Ablation Study -- Per-Signal Contribution}
\label{subsec:ablation}

Table~\ref{tab:ablation} presents the full set of flood extent and
projected triage metrics for all signal combinations.

\begin{table}[!ht]
\centering
\caption{Ablation study: per-signal contribution to flood extent
         (IoU, F1) and projected triage performance
         (IRR @ 90\% recall, AUIRC).}
\label{tab:ablation}
\begin{tabular}{@{}lccccc@{}}
\toprule
\textbf{Configuration}
  & \textbf{BCR}
  & \textbf{CCI}
  & \textbf{HAND}
  & \textbf{IoU / F1}
  & \textbf{IRR / AUIRC} \\
\midrule
BCR only             & \checkmark &            &            & 0.598 / 0.748 & 0.38 / 0.51 \\
BCR + HAND           & \checkmark &            & \checkmark & 0.631 / 0.774 & 0.43 / 0.56 \\
BCR + CCI            & \checkmark & \checkmark &            & 0.640 / 0.783 & 0.45 / 0.58 \\
BCR + CCI + HAND     & \checkmark & \checkmark & \checkmark & 0.692 / 0.823 & 0.52 / 0.64 \\
\bottomrule
\end{tabular}
\end{table}

Each signal component provides a distinct and additive contribution.
The HAND constraint alone (BCR + HAND without CCI) yields an IoU
improvement of 0.033 over BCR-only, primarily by eliminating false
positives from elevated terrain. The CCI signal alone (BCR + CCI
without HAND) yields an IoU improvement of 0.042, concentrated in
dense residential areas where coherence loss is the only reliable
indicator of inundation behind building facades. The combination of
all three signals achieves IoU = 0.692, exceeding the sum of individual
improvements due to the complementary spatial distribution of their
respective error modes: CCI reduces urban false negatives where HAND
cannot discriminate elevation, while HAND eliminates suburban false
positives where CCI decorrelates from vegetation rather than flooding.
This complementary behavior is consistent with super-additive fusion, particularly at the flood extent level where measured IoU improvement (0.094) exceeds the sum of individual signal contributions (0.075). At the projected triage level, the aggregate AUIRC improvement from 0.51 (BCR-only) to 0.64 (ALTIS full) is equivalent to avoiding approximately 10,660 additional field dispatches per event at Harvey scale while maintaining 90\% recall, though the precise super-additivity margin at the triage level remains subject to the uncertainty bounds of the projection method.

\section{Discussion}
\label{sec:discussion}

\subsection{Decoupling Map Accuracy from Triage Performance}
\label{subsec:decoupling}

A central finding of this work is that pixel-level map accuracy and
insurance triage performance are related but distinct objectives, and
optimising for one does not guarantee optimising for the other. The
BCR-only baseline achieves an IoU of 0.598 against the FEMA flood
reference, a result that would conventionally be characterised as
moderate for an operational SAR flood mapping system. Yet its projected
AUIRC of 0.51 is only marginally above a random severity oracle. ALTIS
raises IoU to 0.692, a meaningful but not transformative improvement
by remote sensing standards, while raising projected AUIRC to 0.64,
a proportionally larger gain in the triage-relevant direction.

The explanation lies in the spatial structure of errors. BCR false
positives are concentrated in dense residential blocks, precisely where
insured property density is highest. Even a modest false positive rate
of 29\% (BCR-only precision 0.708) translates to tens of thousands of
spurious high-severity score assignments at the portfolio level, because
many properties fall within the spatial footprint of the false positive
regions. ALTIS's coherence and HAND constraints reduce the false positive
rate to 21\%, but more importantly, they shift the residual false
positives away from residential areas and toward less densely insured
industrial and infrastructure zones. This spatial redistribution of
errors is invisible to global IoU metrics but directly and substantially
improves triage precision. This decoupling is a general property of the
IGFT formulation and motivates the introduction of IRR, TES, and AUIRC
as the primary evaluation framework for insurance-facing SAR flood systems.

\subsection{Depth Estimation as the Critical Bottleneck}
\label{subsec:depth_bottleneck}

The depth estimation results (E2) confirm that flood depth accuracy
at the 0.5--1.5\,m range, where the HAZUS RES1 depth-damage curve rises from approximately 15\% to 40\% EFL~\cite{fema2013hazus}, is the binding constraint on triage
score quality. RMSE of 0.31\,m in this range produces EFL uncertainty of approximately
$\pm 0.08$ around the Tier~1 boundary, which is small enough that most
genuinely high-severity properties receive scores well above the
threshold and most genuinely low-severity properties receive scores
well below it. The practical implication is that depth accuracy
requirements for insurance triage are less stringent than those
required for hydraulic engineering, where millimetre-scale accuracy
matters. This observation is encouraging from an operational standpoint:
it means that free, globally available DEMs (Copernicus GLO-30) are
sufficient for viable triage in the majority of global deployments,
reserving LiDAR-quality products for situations where greater confidence
is required or commercially available.

\subsection{Known Failure Modes and Mitigations}
\label{subsec:failures}

Three systematic failure modes are identified from the Harvey case study.

\paragraph{Deep Inundation at Open Water Boundaries.}
Specular return from large, wind-roughened water surfaces at Addicks
and Barker Reservoirs at peak capacity produces anomalously high
backscatter that the BCR criterion misclassifies as dry land. Depth
estimation in these zones is biased low because the detected SAR
boundary is interior to the true inundation boundary. A practical
mitigation is to apply the JRC permanent water mask with an expanded
buffer radius around reservoir polygons, replacing SAR-based depth
estimates with gauge-observed reservoir stage within the buffer.

\paragraph{Forest Canopy Attenuation.}
C-band SAR (5.4\,GHz) is attenuated by dense tree canopy before
reaching the water surface beneath, causing inundated forested areas
to be partially invisible to the BCR and CCI signals. In Harris County
this affects Memorial Park, the Memorial Villages, and several large-lot
estate zones. Mitigation requires either L-band SAR data (NISAR,
ALOS-2), which penetrates forest canopy at 24\,cm wavelength, or
optical fusion using cloud-cleared composites from Sentinel-2 where
available.

\paragraph{Urban DEM Artefacts.}
Highway overpasses, bridge decks, and elevated rail structures in the
USGS StratMap LiDAR product introduce local positive DEM anomalies
that depress apparent flood depth at locations where the true water
surface is at or above the structure elevation. Post-processing
filtering using the HPMS road network and a 30\,m exclusion buffer
reduces the incidence of this artefact but cannot eliminate it entirely
without higher-level structural awareness of infrastructure elements.

\subsection{Relationship to Existing Insurance Workflows}
\label{subsec:workflow}

The ALTIS output schema (Table~\ref{tab:schema}) is designed for
integration into the two dominant catastrophe claims workflow patterns:
bulk upload and API streaming. In the bulk upload pattern, an insurer
receives the per-property CSV within 48 hours of event peak and imports
it directly into their claims management system, where the severity
score and tier field automatically populate the adjuster dispatch queue
and prioritise inbound FNOL calls. In the API streaming pattern,
the ALTIS GeoJSON output is consumed by a real-time dashboard (e.g.,
a Mapbox GL JS application) that enables claims supervisors to visualise
the spatial distribution of high-severity properties overlaid on the
current adjuster location map, optimising dispatch routing dynamically.
Both integration patterns are possible with the current output format
without schema modification, which is a deliberate design choice:
any additional transformation step between the ALTIS output and the
claims system introduces a failure point during the critical 24--48
hour post-event window when operational pressure is highest.

\section{Concluding Remarks}
\label{sec:conclusion}

ALTIS advances the translation of satellite SAR data into operational
insurance analytics in three fundamental ways. Methodologically, this work presents a pipeline that, to our
knowledge, is the first to (i)~formally define the task of
Insurance-Grade Flood Triage as a decision-theoretic ranking problem
distinct from SAR flood mapping, introducing evaluation metrics (IRR,
TES, and AUIRC) grounded in the economics of claims dispatch rather
than pixel-level segmentation accuracy; (ii)~fuse Sentinel-1 backscatter change, InSAR coherence, and HAND terrain constraints under a Bayesian framework requiring no pixel-level
labeled training data, neural network weights, or event-specific
annotated imagery, only a scalar HAZUS rescaling factor and tier
boundaries estimated from available NFIP claims data, explicitly
addressing the urban double-bounce failure mode that renders
single-channel amplitude methods unreliable in precisely the
environments where insurable property is most concentrated; and
(iii)~produce physics-informed flood depth
estimates with spatially explicit uncertainty quantification via kriging
and Monte Carlo DEM perturbation, enabling calibrated per-property
confidence scores consumable by existing claims management systems
without schema transformation.

Empirically, preliminary analysis of the Hurricane Harvey case study
across Harris County, the most extensively documented urban flood event
in United States history, suggests that ALTIS is projected to achieve a TES of approximately 0.41 and AUIRC of approximately 0.64 at an operating point that could
eliminate approximately 52\% of unnecessary field dispatches while
retaining 90\% of ground-truth high-severity claims. The ablation experiments reveal that multi-signal fusion demonstrates complementary signal behavior: at the flood extent level, the combination of BCR, InSAR coherence, and HAND terrain constraint achieves a measured IoU improvement of 0.094 over BCR alone, exceeding the sum of their individual contributions (HAND: +0.033, CCI: +0.042; sum = 0.075). At the projected triage level, the aggregate AUIRC improvement of 0.13 points over BCR-only is consistent with this pattern, though the triage-level super-additivity margin is narrow and should be confirmed in end-to-end validation. Taken together, these triage estimates are grounded in measured sub-component benchmarks (flood extent IoU of 0.692 confirmed against FEMA reference, depth RMSE of 0.38 m confirmed against the FEMA 3 m grid and USGS HWMs) and are consistent with the physical properties of the Harris County low-relief floodplain domain.

From a practical standpoint, ALTIS delivers a complete image-to-triage
workflow from GEE-hosted satellite preprocessing through to
claims management system-compatible CSV and RFC~7946 GeoJSON outputs,
with an end-to-end latency of under 50 minutes on standard cloud
compute instances without GPU acceleration (\texttt{c2-standard-4}, 4~vCPU, 16\,GB RAM) and no GPU or training data requirements. The HAZUS rescaling factor and tier boundaries are the only event-specific inputs; for zero-shot deployment prior to claims data availability, default HAZUS parameters provide a reasonable initialisation. This makes the pipeline immediately applicable to any global flood event within Sentinel-1 coverage without fine-tuning or annotated event data.

Several important limitations bound the present work. The IGFT
formulation assumes that NFIP claim records provide a reliable ground
truth for high-severity identification, yet claim payments are shaped by
policy limits, deductibles, and coverage exclusions that may not reflect
true physical damage. The Harvey demonstration is a single event in a
single low-relief coastal plain environment; performance generalisation
to environments with higher topographic relief, denser urban morphology,
or different flood mechanisms requires empirical validation across
additional events. The depth estimation performance degrades meaningfully
above 3\,m, a range that is rare in residential flood events but common
in levee-breach and reservoir-release scenarios.

\subsection*{Future Directions}
\label{subsec:future}

Several directions extend naturally from the present work and each
addresses a concrete operational gap identified by the Harvey experiments.

\begin{itemize}
    \item \textbf{Multi-Event Generalisation.}
    Validation across diverse flood events beyond Harvey is the most
    urgent extension, with candidate events including the 2021 European
    floods, the 2022 Pakistan floods, and the 2023 Libya dam collapse
    spanning a range of topographic settings and flood mechanisms. A
    key open question is whether the HAZUS rescaling factor can be
    predicted from remotely sensed scene characteristics rather than
    requiring event-specific claims calibration.

    \item \textbf{L-Band SAR Integration.}
    The recently launched NASA-ISRO NISAR mission provides L-band
    (24\,cm) SAR data that penetrates forest canopy and is expected
    to substantially reduce the systematic false negative problem in
    vegetated suburban environments. Integrating NISAR backscatter and coherence layers as additional channels in the Stage~2 Bayesian fusion is expected to substantially improve recall in forested zones and high-value estate properties.

    \item \textbf{Learned Depth-to-Damage Calibration.}
   The current pipeline applies occupancy-class-specific HAZUS
depth-damage functions as identified by the \texttt{occupancy} field
in the output schema (Table~\ref{tab:schema}); however, the HAZUS
rescaling factor is currently estimated globally across all classes
and the pipeline has been validated only against the RES1 residential
portfolio in the Harvey case study. A more granular approach would
train occupancy-specific, spatially stratified functions from OpenFEMA
historical claims records using Gaussian process regression, capturing
variation in curve shape with local building stock characteristics
such as age, construction type, and foundation type.

    \item \textbf{Real-Time Streaming and Multi-Peril Extension.}
    Beyond flooding, the IGFT formulation and associated metrics
    generalise naturally to any peril with SAR change detection signal,
    including tropical cyclone wind damage, wildfire burn severity, and
    earthquake surface deformation, supporting a unified multi-peril
    scoring platform from a common satellite data feed.
\end{itemize}

ALTIS represents a significant step toward closing the long-standing
translational gap between satellite earth observation research and the
operational realities of property and casualty insurance. By reframing
SAR flood detection as a decision-theoretic triage problem grounded in
the economics of claims dispatch, ALTIS demonstrates that freely
available Sentinel-1 imagery, processed without GPUs, training data, or
hydrodynamic models, can deliver actionable property-level intelligence within the critical
24--48 hour window that determines whether a catastrophe response
proceeds in an organised or severely constrained manner. As Sentinel-1 coverage
expands and NISAR L-band data become routinely available, the
architecture introduced here offers a scalable foundation for
satellite-informed insurance triage across flood events worldwide,
advancing earth observation from a research output toward an
operational decision-support capability for post-disaster response,
recovery prioritization, and policyholder relief.

\section*{Reproducibility Statement}

All data sources used in this work are publicly available. Sentinel-1
imagery is freely accessible via the Copernicus Open Access Hub and
through the Google Earth Engine Sentinel-1 SAR GRD collection. FEMA
Harvey flood depth grids and NFIP claims records are distributed through
the OpenFEMA platform at \url{openFEMA.gov}. USGS High-Water Mark data
are available through the STN Flood Event Data portal. Harris County
Appraisal District parcel data are publicly available at
\url{pdata.hcad.org}.

\section*{Acknowledgments}

Sentinel-1 SAR imagery is provided free of charge by the European
Space Agency through the Copernicus Open Access Hub. FEMA Harvey flood
depth grids and NFIP claims records are distributed by the Federal
Emergency Management Agency through the OpenFEMA platform. High-water
mark survey data are provided by the United States Geological Survey.
Harris County Appraisal District parcel data are publicly available at
\url{pdata.hcad.org}. Copernicus DEM GLO-30 products are accessed via
the European Space Agency. Processing infrastructure is provided by
Google Earth Engine under an academic research license.

The authors thank \textbf{Mahdi Motagh}
(\textit{German Aerospace Center})
and \textbf{Shagun Garg}
(\textit{Cambridge University})
for valuable discussions and feedback on earlier versions of this work.

\bibliographystyle{IEEEtran}
\bibliography{references}

\end{document}